\documentclass[runningheads]{llncs}
\usepackage[T1]{fontenc}
\usepackage{graphicx}
\usepackage{amsmath}
\usepackage{amssymb}
\usepackage{hyperref}
\usepackage{subfigure}
\usepackage{booktabs}
\usepackage{multirow}
\usepackage{threeparttable}

\usepackage{marvosym}
\newcommand{\corr}{(\Letter)}
\begin{document}
\title{Learning Accurate, Efficient, and Interpretable MLPs on Multiplex Graphs via Node-wise Multi-View Ensemble Distillation}
\titlerunning{Distilling Multiplex GNNs into MLPs}
%

\author{Yunhui~Liu\inst{1} \and
Zhen~Tao\inst{1} \and
Xiang~Zhao\inst{2} \and
Jianhua~Zhao\inst{1} \and
Tao~Zheng\inst{1} \and
Tieke~He\inst{1} \corr}

\authorrunning{Y. Liu et al.}

\institute{State Key Laboratory for Novel Software Technology, Nanjing University, Nanjing, China \email{\{lyhcloudy1225,shchenen,hetieke\}@gmail.com}
\and
Laboratory for Big Data and Decision, National University of Defense Technology, Hunan, China
\email{xiangzhao@nudt.edu.cn}}


\maketitle              
\begin{abstract}

Multiplex graphs, with multiple edge types (graph views) among common nodes, provide richer structural semantics and better modeling capabilities. Multiplex Graph Neural Networks (MGNNs), typically comprising view-specific GNNs and a multi-view integration layer, have achieved advanced performance in various downstream tasks. However, their reliance on neighborhood aggregation poses challenges for deployment in latency-sensitive applications. 
Motivated by recent GNN-to-MLP knowledge distillation frameworks, we propose Multiplex Graph-Free Neural Networks (MGFNN and MGFNN+) to combine MGNNs' superior performance and MLPs' efficient inference via knowledge distillation. 
MGFNN directly trains student MLPs with node features as input and soft labels from teacher MGNNs as targets. MGFNN+ further employs a low-rank approximation-based reparameterization to learn node-wise coefficients, enabling adaptive knowledge ensemble from each view-specific GNN. This node-wise multi-view ensemble distillation strategy allows student MLPs to learn more informative multiplex semantic knowledge for different nodes.
Experiments show that MGFNNs achieve average accuracy improvements of about 10\% over vanilla MLPs and perform comparably or even better to teacher MGNNs (accurate); MGFNNs achieve a 35.40×-89.14× speedup in inference over MGNNs (efficient); MGFNN+ adaptively assigns different coefficients for multi-view ensemble distillation regarding different nodes (interpretable).
\end{abstract}
\section{Introduction}
Multiplex graphs, characterized by the presence of multiple edge types (graph views) among a common set of nodes, provide a more realistic representation of complex systems encapsulating multiple structural relations among nodes in the real world. Due to their strong modeling abilities, multiplex graphs have gained widespread popularity and application across various domains such as paper classification in academic networks \cite{HAN}, item recommendation in e-commerce networks \cite{MGNN}, fraud detection in financial networks \cite{CARE-GNN}, and molecular property prediction in biological networks \cite{PAMNet}.

Recently, multiplex graph neural networks (MGNNs) have demonstrated significant effectiveness in mining multiplex graphs \cite{RGCN,HAN,HPN,MHGCN}. MGNNs generate node embeddings through recursively aggregating messages from neighboring nodes across various relation types. By employing this relation-aware message-passing mechanism, MGNNs effectively model the diverse structural semantics inherent in multiplex graphs, achieving state-of-the-art performance. 
However, this message-passing mechanism inevitably makes deploying MGNNs for real-world industrial applications challenging, particularly in environments characterized by large-scale data, limited memory, and high sensitivity to latency, such as real-time financial fraud detection. 

\begin{figure}[h]
    \centering
    \subfigure[\# Nodes Fetched vs. \# Layers]{\includegraphics[width=0.45\linewidth]{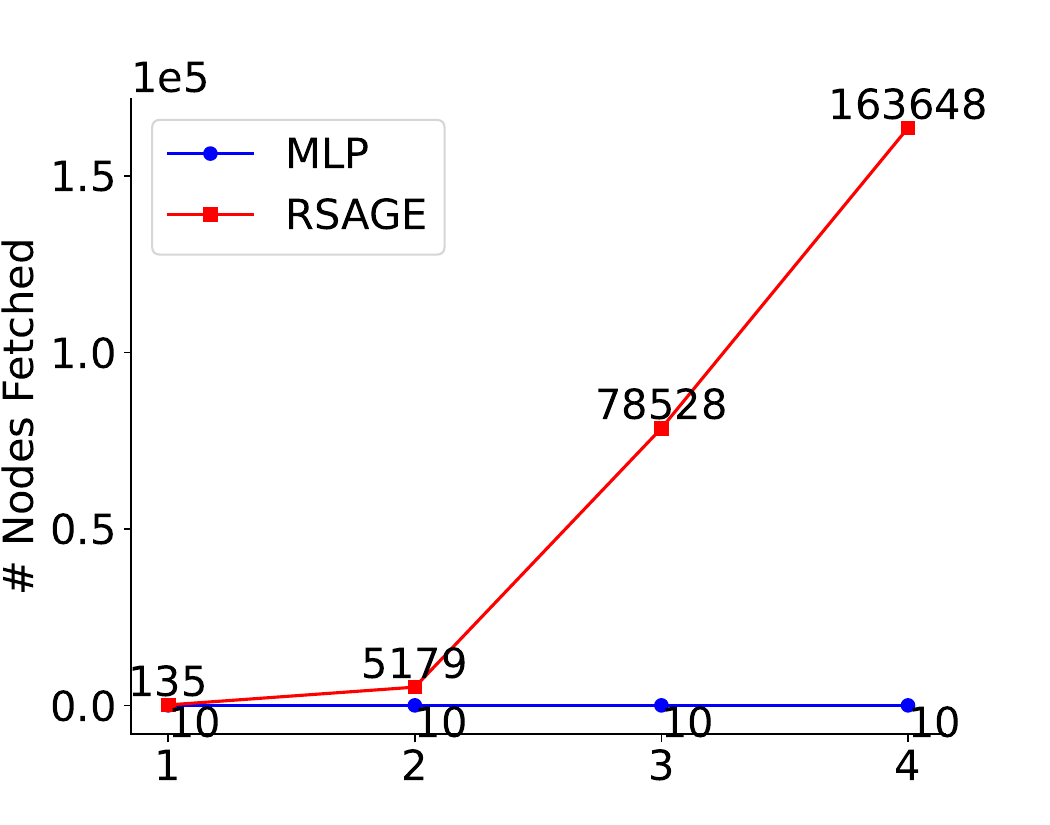}}
    \subfigure[Inference Time vs. \# Layers]{\includegraphics[width=0.45\linewidth]{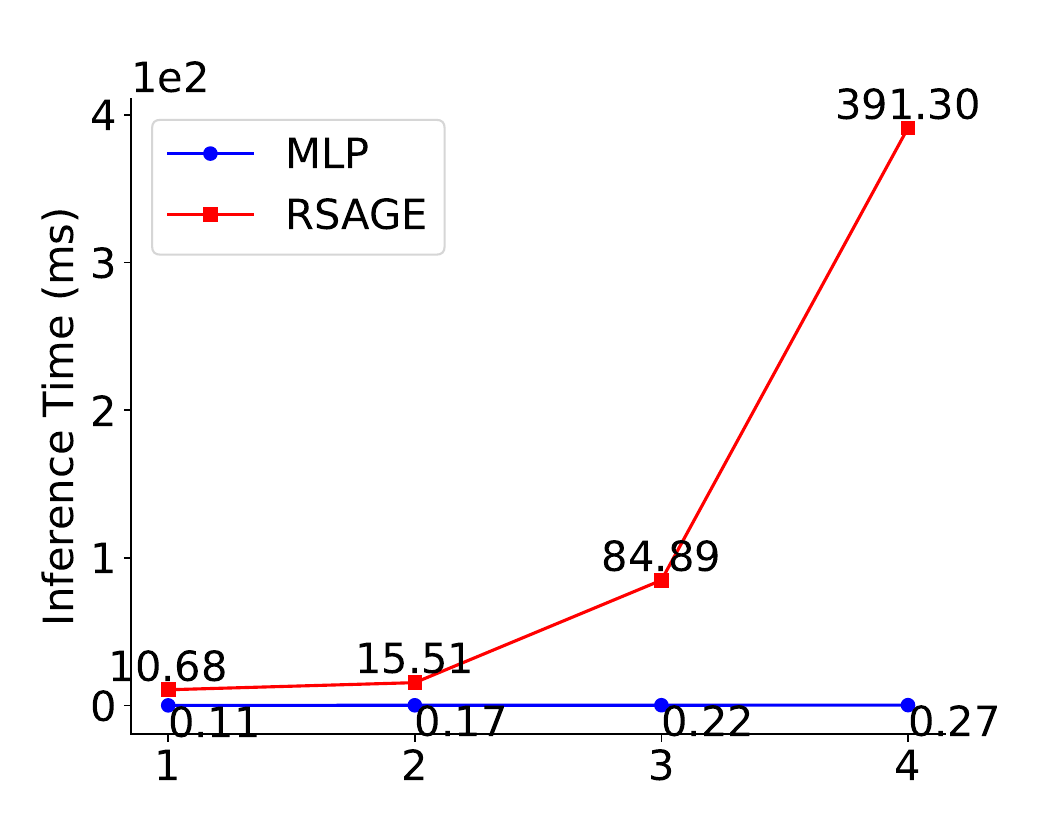}}
    \caption{The number of nodes fetched and inference time of MGNNs are both magnitudes more than MLPs and grow exponentially with the number of layers. (a) The total number of nodes fetched for inference. (b) The total inference time. (Inductive inference for $10$ random nodes on MAG.)}\label{Fig: Neighborhood Fetching and Inference Time}
\end{figure}

The primary obstacle is MGNNs' dependence on graph structure during inference. Specifically, inference for a target node requires fetching the features of numerous neighboring nodes based on the graph topology, causing the number of nodes fetched and inference times to grow exponentially with the number of MGNN layers \cite{GLNN}. Furthermore, MGNNs process different edge types (graph views) independently, resulting in computational costs that increase in proportion to the number of edge types in multiplex graphs. As shown in Figure \ref{Fig: Neighborhood Fetching and Inference Time}, adding more MGNN layers exponentially increases the number of nodes fetched and the inference time. Conversely, MLPs exhibit a much smaller and linearly growing inference time, as they only process node features. However, this lack of structural information often limits MLPs' performance compared to MGNNs.

Given the trade-offs between efficiency and accuracy, several recent studies \cite{GLNN,NOSMOG,KRD,HG2M} have proposed knowledge distillation frameworks that transfer knowledge from GNNs to MLPs. These frameworks facilitate significantly faster inference while maintaining competitive performance compared to teacher GNNs. However, existing research has primarily focused on homogeneous graphs, and distilling MGNNs into MLPs for multiplex graphs has yet to be explored. Multiplex graphs, which capture diverse types of information that represent complex semantic relationships between nodes, present a challenge for current GNN-to-MLP methods, as they are insufficient for handling such multiplexity. Therefore, we pose the following question: \emph{Can we bridge the gap between MLPs and MGNNs to enable extremely efficient inference while effectively distilling multiplex semantics?}

\textbf{Present Work.} 
In this paper, we propose Multiplex Graph-Free Neural Networks: MGFNN and MGFNN+, which combine the superior accuracy performance of MGNNs with the efficient inference capabilities of MLPs. MGFNN directly employs knowledge distillation \cite{KD} to transfer knowledge learned from teacher MGNNs to student MLPs using soft labels.
We then empirically show that relying solely on soft labels from teacher MGNNs may lead to suboptimal knowledge distillation, as it does not fully capture the rich, detailed semantic knowledge provided by each view-specific GNN. To address this issue, MGFNN+ introduces a multi-view ensemble distillation strategy, which adaptively injects multiplex semantic knowledge from view-specific GNNs into student MLPs. It introduces a low-rank approximation-based reparameterization method to learn node-wise ensemble coefficients, improving adaptability and reducing parameter cost.
Experiments conducted on six real-world multiplex graph datasets validate the effectiveness and efficiency of MGFNNs. In terms of performance, under a production setting encompassing both transductive and inductive predictions, MGFNNs achieve average accuracy improvements of about 10\% over vanilla MLPs and perform comparably to teacher MGNNs on 5/6 datasets (see Section \ref{Sec: Accuracy}). 
Regarding efficiency, MGFNNs deliver an inference speedup ranging from 35.40× to 89.14× compared to teacher MGNNs (see Section \ref{Sec: Efficiency}). 
Additionally, MGFNN+ can learn different ensemble coefficients to distill multiplex semantic knowledge for different nodes interpretably (see Section \ref{Sec: Interpretability}).
These results suggest that our MGFNNs are a better choice for accurate and fast inference in multiplex graph learning, especially in latency-sensitive applications. 
The code is available at \url{https://github.com/Cloudy1225/MGFNN}. 
In conclusion, our contributions can be summarized as follows:
\begin{itemize}
    \item We are the first to integrate the superior performance of MGNNs with the efficient inference of MLPs through knowledge distillation.

    \item We propose a node-wise multi-view ensemble distillation strategy to inject more informative multiplex semantic knowledge into student MLPs.

    \item To reduce the heavy burden of learning node-wise ensemble coefficients, we introduce a low-rank approximation-based reparameterization method for decomposing the ensemble coefficient matrix.

    \item Experiments on six datasets show MGFNNs that MGFNNs achieve competitive performance with MGNNs, significantly outperform vanilla MLPs, and provide 35.40×-89.14× faster inference compared to teacher MGNNs.
\end{itemize}

\section{Related Work}
\subsection{Multiplex Graph Neural Networks}
Multiplex Graph Neural Networks (MGNNs) utilize relation-aware message-passing to capture complex relationships and diverse semantics within multiplex graphs. For instance, RGCN \cite{RGCN} initially embeds each graph view separately using view-specific GCNs \cite{GCN}, and then employs an average pooling on these multiple node embeddings to generate final embeddings. Note that the graph convolutions in RGCN can be substituted with other classical GNNs, such as SAGE \cite{SAGE}, GAT \cite{GAT}, or SGC \cite{SGC}. NARS \cite{NARS} uses SGC to generate node embeddings for each graph view and combines them through a learnable 1D convolution. Additionally, HAN \cite{HAN} applies GAT to encode each view and proposes a semantic attention mechanism to aggregate resulted node embeddings. HPN \cite{HPN} designs a semantic propagation mechanism to reduce semantic confusion, also employing the attention-based multi-view fusion mechanism. Despite these advancements, the inherent structural dependency of MGNNs presents challenges for deployment in latency-sensitive applications that require rapid inference.

\subsection{GNN-to-MLP Knowledge Distillation}
In response to latency concerns, recent studies have sought to bridge the gaps between powerful GNNs and lightweight MLPs through knowledge distillation \cite{KD}. A pioneering effort, GLNN \cite{GLNN}, directly transfers knowledge from teacher GNNs to vanilla MLPs by imposing Kullback-Leibler divergence between their logits. To distill reliable knowledge, KRD \cite{KRD} develops a reliable sampling strategy while RKD-MLP \cite{RKD-MLP} utilizes a meta-policy to filter out unreliable soft labels. FF-G2M \cite{FF-G2M} employs both low- and high-frequency components in the spectral domain for comprehensive knowledge distillation. NOSMOG \cite{NOSMOG} enhances the performance and robustness of student MLPs by introducing positional features, representational similarity distillation, and adversarial feature augmentation. VQGraph \cite{VQGraph} learns a powerful new graph representation space by directly labeling nodes according to their diverse local structures for distillation. AdaGMLP \cite{AdaGMLP} addresses the challenges of insufficient training data and incomplete test data through ensemble learning and node alignment, while MTAAM \cite{MTAAM} amalgamates various GNNs into a super teacher. 
LLP \cite{LLP} and MUGSI \cite{MuGSI} propose GNN-to-MLP frameworks tailored for link prediction and graph classification, and LightHGNN \cite{LightHGNN} extends this methodology to hypergraphs. However, the distillation of MGNNs into MLPs for multiplex graphs remains unexplored.

\section{Preliminaries}
\subsection{Problem Definition}
\subsubsection{Multiplex Graph.}
A multiplex graph is denoted by $\mathcal{G} = \{ \mathcal{G}_1, \mathcal{G}_2, \dots, \mathcal{G}_r \}$, where $\mathcal{G}_i=\{ \mathcal{V}, \mathcal{E}_i, \boldsymbol{A}_i, \boldsymbol{X} \}$ is the $i$-th view corresponding to the $i$-th view. For each view $\mathcal{G}_i$, $\mathcal{V}$ and $\mathcal{E}_i$ denote the node set and edge set, respectively; $\boldsymbol{A}_i \in \{0,1\}^{n \times n}$ is the associated adjacency matrix, and $\boldsymbol{X} \in \mathbb{R}^{n \times d}$ is the shared feature matrix across all views. 

\subsubsection{Node Classification.}
Considering node classification, we have the label matrix $\boldsymbol{Y} \in \mathbb{R}^{n \times k}$, where row $\boldsymbol{y}_v$ is a $k$-dimensional one-hot vector for node $v \in \mathcal{V}$. We use the superscript $L$ and $U$ to divide $\mathcal{V}$ into labeled $(\mathcal{V}^L, \boldsymbol{X}^L, \boldsymbol{Y}^L)$ and unlabeled parts $(\mathcal{V}^U, \boldsymbol{X}^U, \boldsymbol{Y}^U)$. Our objective is to predict $\boldsymbol{Y}^U$, with $\boldsymbol{Y}^L$ available.

\subsection{Multiplex Graph Neural Networks}
GNNs usually utilize the message-passing mechanism to propagate and aggregate neighborhood information. For multiplex graphs, a typical framework involves separately embedding each view using view-specific GNNs and subsequently applying a multi-view integration function to these multiple node embeddings to generate final embeddings. Formally, the node embedding $\boldsymbol{h}_{v}^{(l)}$ at $l$-th layer of MGNNs can be written as
\begin{equation}
\begin{aligned}
    \boldsymbol{h}_{v}^{(l)} 
    &= \underset{\forall i \in \{1,2,\dots,r\}} {\textbf{Integrate}} \left( \textbf{GNN}_i\left(
    \left\{ \boldsymbol{h}_{u}^{(l-1)}|\forall u \in \mathcal{N}_i(v) \right\}; \boldsymbol{h}_{v}^{(l - 1)}\right)  \right)\\
    &= \underset{\forall i \in \{1,2,\dots,r\}} {\textbf{Integrate}} \left( \underset{\forall u \in \mathcal{N}_i(v)} {\textbf{Aggregate}} \left(\textbf{Propagate}\left(\boldsymbol{h}_{u}^{(l-1)}; \boldsymbol{h}_{v}^{(l - 1)}\right)\right) \right). \label{EQ: MGNN}
\end{aligned}
\end{equation}
Here, $\mathcal{N}_i(v)$ denotes the neighbor set of node $v$ corresponding to the $i$-th graph view (edge type). The parameters of the $\textbf{Propagate}\left(\cdot\right)$ and $\textbf{Aggregate}\left(\cdot\right)$ functions depend on view-specific $\textbf{GNN}_i$. The $\textbf{Integrate}\left(\cdot\right)$ function may utilize either simple average pooling or attention-based fusion. With these designs, MGNNs can effectively capture diverse structural semantics in multiplex graphs. A successful instantiation of this framework is HAN \cite{HAN}, which leverages view-specific GATs \cite{GAT} to embed each view and employs attention-readout on node embeddings of all views to generate final node embeddings.

\section{Methodology}
\subsection{MGFNN: Multiplex Graph-Free Neural Networks}
Similar to GLNN \cite{GLNN}, the key idea of MGFNN is simple yet effective: teaching vanilla MLPs to master multiplex graph-structured knowledge via distillation \cite{KD}. 
Specifically, we generate soft targets $\boldsymbol{z}_v$ for each node $v$ using well-trained teacher MGNNs. Then we train student MLPs supervised by both true labels $\boldsymbol{y}_v$ and $\boldsymbol{z}_v$. 
The objective is as Eq. \eqref{Eq: MGFNN}, with $\lambda$ being a weight parameter, ${CE}$ being the Cross-Entropy loss between student predictions $\hat{\boldsymbol{y}}_v$ and $\boldsymbol{y}_v$, ${KL}$ being the Kullback-Leibler divergence loss between $\hat{\boldsymbol{y}}_v$ and $\boldsymbol{z}_v$.
\begin{equation}
    \mathcal{L} = \lambda\mathcal{L}_{CE} + (1-\lambda)\mathcal{L}_{KL} = 
    \lambda \sum_{v \in \mathcal{V}^L} {CE}\left( \hat{\boldsymbol{y}}_v, \boldsymbol{y}_v \right) + (1-\lambda) \sum_{v \in \mathcal{V}} {KL}\left( \hat{\boldsymbol{y}}_v, \boldsymbol{z}_v \right). \label{Eq: MGFNN}
\end{equation}
The model after distillation, i.e., MGFNN, is essentially an MLP. Therefore, during inference, MGFNN has no dependency on the multiplex graph structure, allowing it to perform as efficiently as vanilla MLPs. Additionally, through distillation, MGFNN parameters are optimized to predict and generalize comparably to MGNNs, with the added benefit of faster inference and easier deployment. 

\subsection{MGFNN+: Node-wise Multi-View Ensemble Distillation}
However, MGFNN uses only the final outputs $\boldsymbol{z}_v$ from MGNNs as soft targets for knowledge distillation, which limits its ability to fully leverage the diverse semantic knowledge captured by each view-specific GNN in MGNNs. According to Eq. \eqref{EQ: MGNN}, the soft targets $\boldsymbol{z}_v$ used in Eq. \eqref{Eq: MGFNN} is essentially an integration of the predictions $\boldsymbol{z}_v^i$ from each view-specific GNN, i.e., $\boldsymbol{z}_v = \sum_{i=1}^r \alpha^i \boldsymbol{z}_v^i$, where $\alpha_i$ is the view-wise fusion weight; for instance, $\alpha^i = 1/r$ when applying average fusion. 
We argue that the view-wise fusion weights are not optimally adaptive for every node and thus $\boldsymbol{z}_v$ loses much critical and accurate information in $\left\{ \boldsymbol{z}_v^i \right\}_{i=1}^r$.

\begin{figure}[h]
    \centering
    \subfigure[ACM]{\includegraphics[width=0.325\linewidth]{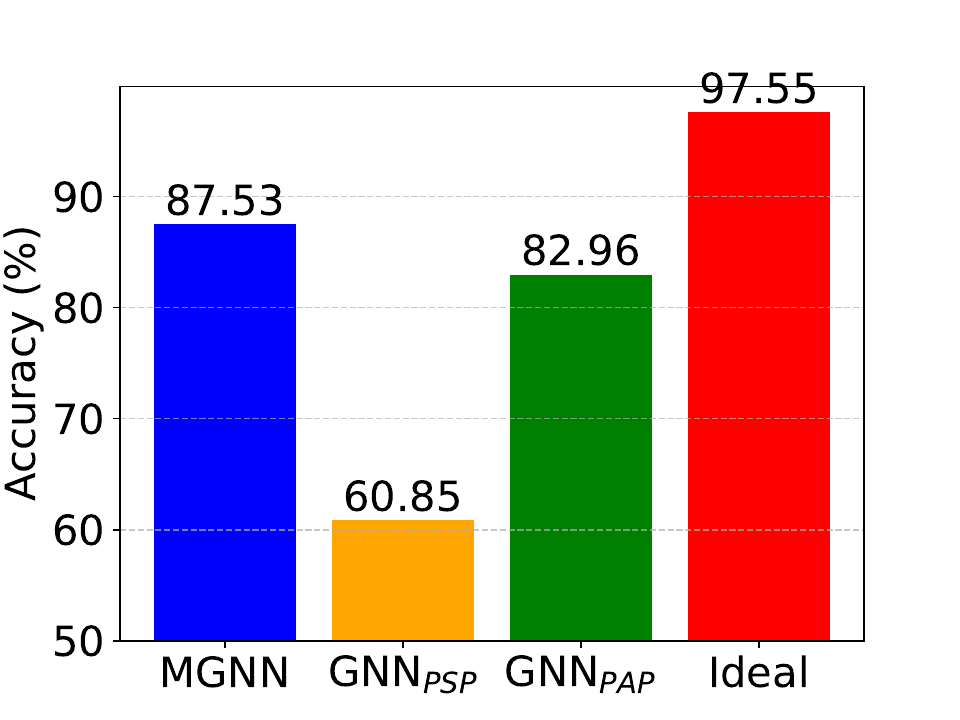}}  
    \subfigure[IMDB]{\includegraphics[width=0.325\linewidth]{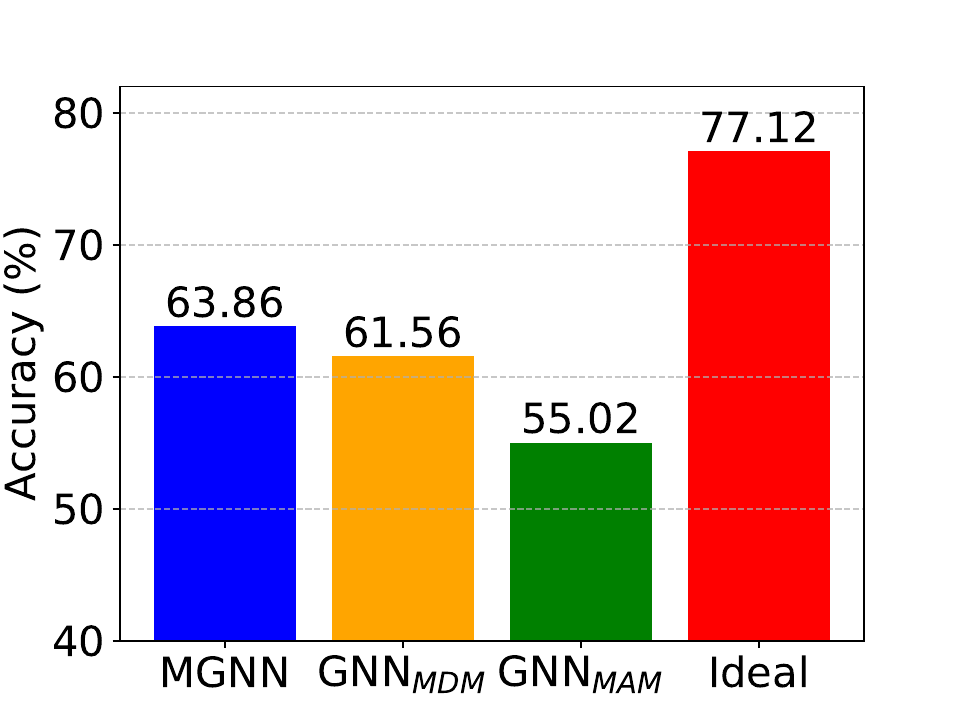}}
    \subfigure[MAG]{\includegraphics[width=0.325\linewidth]{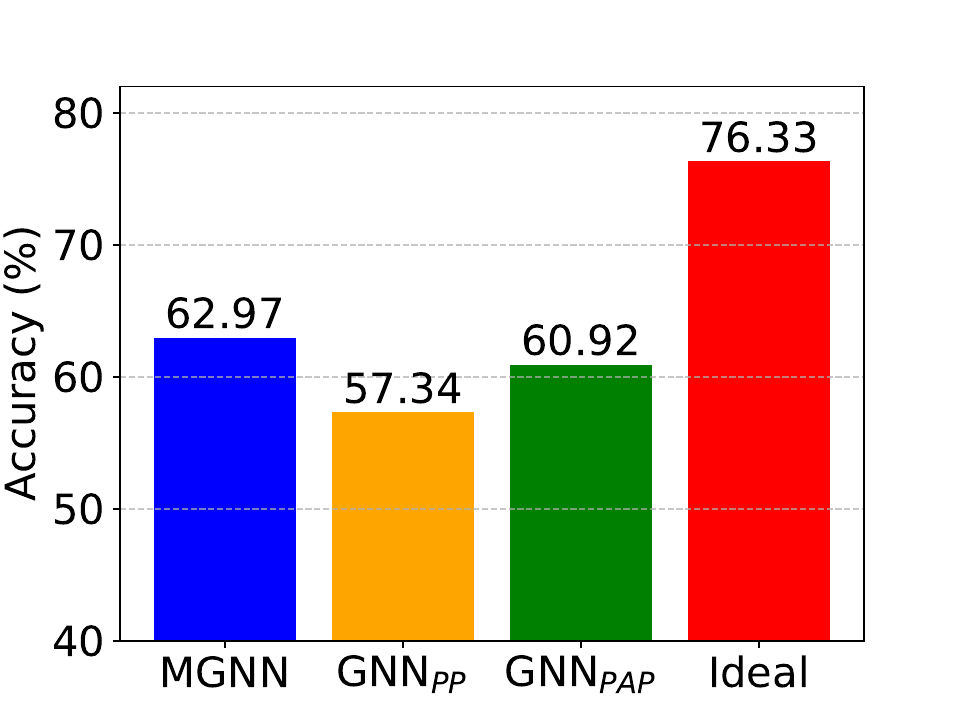}}
    \vspace{-0.2cm}
    \caption{Classification accuracy of MGNN, each view-specific GNN, and the ideal ensemble classifier on ACM, IMDB, and MAG.}
    \label{Fig: Empirical Study}
\end{figure}

\textbf{Empirical Analysis.}\quad
To verify our argument, we conduct an exploratory experiment from an oracle perspective. Specifically, we first evaluate the performance of each view-specific GNN in a well-trained MGNN (instantiated as RSAGE). We then compute the performance of an ideal ensemble classifier, which can make correct predictions as long as one of the view-specific GNNs or the well-trained MGNN predicts correctly. Figure \ref{Fig: Empirical Study} presents the results across three datasets, revealing two key observations: 1) The superior performance of the MGNN compared to each view-specific GNN shows that MGNN effectively integrates outputs from view-specific GNNs to some extent. 2) However, the significantly lower performance of the MGNN than the ideal classifier indicates that much critical and accurate information from the individual GNNs is lost when merging outputs into $\boldsymbol{z}_v$. 
This suggests that relying solely on the final integrated predictions $\boldsymbol{z}_v$ may lead to suboptimal knowledge distillation, as it does not fully capture the rich, detailed semantic knowledge provided by each view-specific GNN.

\subsubsection{Node-wise Multi-View Ensemble Distillation}
The empirical study underscores the need for more sophisticated methods to leverage the full potential of multiplex semantic knowledge in view-specific GNNs. One intuitive solution is multi-view ensemble distillation, which treats all view-specific GNNs as additional teacher models to supervise student MLPs. This can be simply implemented by imposing additional KL loss items between soft labels $\left\{ \boldsymbol{z}_v^i \right\}_{i=1}^r$ and $\boldsymbol{z}_v$. Given the varying importance and accuracy of the soft labels from each GNN, it is essential to assign different weights during ensemble distillation. For simplicity, we denote the MGNN as a \emph{Whole} view-specific GNN and denote its output $\boldsymbol{z}_v$ as $\boldsymbol{z}_v^{r+1}$. Then, the overall KL loss can be formulated as follows:
\begin{equation}
    \mathcal{L}_{KL} = \sum_{v \in \mathcal{V}} \sum_{i=1}^{r+1} c^i {KL} 
    \left( \hat{\boldsymbol{y}}_v, \boldsymbol{z}_v^i \right), \label{Eq: View-wise MVED}
\end{equation}
where $c^i > 0$ is the coefficient to balance the contributions of each view-specific GNN and $\sum_{i=1}^{r+1} c^i = 1$. These coefficients may be uniformly set to $1/(r+1)$, defined as $(r+1)$-dimensional learnable parameters, or computed adaptively using attention \cite{MTAAM,MSKD} or gradient \cite{AEKD}. However, since different nodes exhibit distinct local structural patterns across various views, these view-wise coefficients only reflect the global importance of each view without but make no specific discrimination for individual nodes during ensemble distillation. These coarse-grained ensembles still do not fully exploit the multiplex semantic knowledge captured by each view-specific GNN, leading to suboptimal results, as seen in Table \ref{Tab: Ablation Study}.
To address this limitation, a more comprehensive and flexible way is to learn appropriate \emph{node-wise} ensemble coefficients specific to different nodes to accommodate their local structural patterns and distill fine-grained semantic knowledge. Thus, the overall KL loss can be reformulated as:
\begin{equation}
    \mathcal{L}_{KL} = \sum_{v \in \mathcal{V}} \sum_{i=1}^{r+1} c_v^i {KL} 
    \left( \hat{\boldsymbol{y}}_v, \boldsymbol{z}_v^i \right). \label{Eq: Node-wise MVED}
\end{equation}
Let $\boldsymbol{C}=[c_v^i] \in \mathbb{R}_+^{n \times (r+1)}$ be the ensemble coefficient matrix for all nodes. The sum of each row of $\boldsymbol{C}$ is constrained to 1, which can be simply achieved using \emph{softmax} normalization. We could treat $\boldsymbol{C}$ as learnable parameters or hyperparameters. However, several concerns regarding this scheme should be noted: 1) The size of $\boldsymbol{C}$ is positively proportional to the number of nodes. As $n$ increases, this inevitably results in a significant number of parameters to learn. 2) Eq. \eqref{Eq: Node-wise MVED} shows that only the gradients from $\hat{\boldsymbol{y}}_v$ can update the ensemble parameters corresponding to node $v$, i.e., the $v$-th row of $\boldsymbol{C}$. This suggests that $\boldsymbol{C}$ is hard to be efficiently optimized. 3) Directly treating $\boldsymbol{C}$ as learnable parameters does not explicitly use the input node features, which may offer valuable information beneficial for learning ensemble coefficients.

\textbf{Learning $\boldsymbol{C}$ via Low-Rank Reparameterization.}\quad
To address the above issues, we leverage the concept of low-rank matrix factorization and propose a separable reparameterization strategy to indirectly learn $\boldsymbol{C}$. Specifically, we decompose $\boldsymbol{C}$ as $\boldsymbol{C} = \boldsymbol{S}\boldsymbol{T}$, where $\mathbf{U} \in \mathbb{R}^{n \times m}$ and $\boldsymbol{T}\in \mathbb{R}^{m \times (r+1)}$ are trainable parameter matrices. 
As can be easily observed, $\boldsymbol{C}_{v:} = \boldsymbol{S}_{v:}\boldsymbol{T}=\sum_{j=1}^m \boldsymbol{S}_{v,j} \boldsymbol{T}_{j:}$. This means that each row of $\boldsymbol{T}$, i.e., $\{t_i\}_{i=1}^{r+1}$, parameterizes a globally shared view-wise weight-assigner. 
Therefore, $\boldsymbol{T}$ represents a set of base view-wise weight-assigners, while $\boldsymbol{S}_{v:}$ is the weights to combine these base assigners for node $v$. 
In other words, the ensemble coefficients $\boldsymbol{C}_{v:}$ specific to $v$ can be obtained by a weighted combination of the base view-wise weight-assigners in $\boldsymbol{T}$. Furthermore, $\boldsymbol{S}$ is a node-dependent trainable matrix that establishes a close link between $\boldsymbol{C}$ and $\boldsymbol{X}$.

Since $\boldsymbol{T}$ is node-agnostic, it can be directly trained as learned parameters. In contrast, since we treat $\boldsymbol{S}$ as node dependent, we extract the output $\boldsymbol{H} \in \mathbb{R}^{n \times h}$ from the last hidden layer of the student MLPs and apply a simple yet effective nonlinear transformation:
\begin{equation}
    \boldsymbol{S} = \tanh \left(\boldsymbol{H} \boldsymbol{W} \right),
\end{equation}
where $\boldsymbol{W} \in \mathbb{R}^{h \times m}$ is the learnable weight matrix and $\tanh(\cdot)$ is the activation function. Using the low-rank reparameterization strategy, the ensemble coefficients $\boldsymbol{C}$ are computed by two matrix multiplications: $\boldsymbol{C} = \boldsymbol{S}\boldsymbol{T} = \tanh \left(\boldsymbol{H} \boldsymbol{W} \right) \boldsymbol{T}$. Given that $\boldsymbol{H} \in \mathbb{R}^{n \times h}$, $\boldsymbol{W} \in \mathbb{R}^{h \times m}$, and $\boldsymbol{T}\in \mathbb{R}^{m \times (r+1)}$, the computational complexity of the low-rank reparameterization is $\mathcal{O}\left(n \times m \times (h+r+1)\right)$.
Additionally, to prevent the learned $\boldsymbol{C}$ from assigning excessively large weight, e.g., $c_v^i > 0.999$, to the $i$-th view-specific teacher for node $v$, which may hinder distilling diverse semantic knowledge, we incorporate a mean entropy maximization regularization. Denote the average coefficients across all views by
$\overline{\boldsymbol{c}} = \frac{1}{n(r+1)} \sum_{v \in \mathcal{V}}\sum_{i=1}^{r+1} c_v^i$. The regularization term simply seeks to maximize the entropy of $\overline{\boldsymbol{c}}$, i.e., $\mathcal{H}(\overline{\boldsymbol{c}}) = -\sum_{i=1}^{r+1} \overline{\boldsymbol{c}}_i \log \overline{\boldsymbol{c}}_i$. Thus, the final loss of MGFNN+ is formulated as:
\begin{equation}
    \mathcal{L} = \lambda \sum_{v \in \mathcal{V}^L} {CE}\left( \hat{\boldsymbol{y}}_v, \boldsymbol{y}_v \right) + (1-\lambda) \left( \sum_{v \in \mathcal{V}} \sum_{i=1}^{r+1} c_v^i {KL} 
    \left( \hat{\boldsymbol{y}}_v, \boldsymbol{z}_v^i \right) - \gamma \mathcal{H}(\overline{\boldsymbol{c}}) \right).\label{Eq: MGFNN+}
\end{equation}

The advantages of the low-rank approximation-based reparameterization are clear and significant. 
First, it reduces the parameter complexity of $\boldsymbol{C}$ from $\mathcal{O}\left(n \times (r+1)\right)$ to $\mathcal{O}\left( (r+1)\times m + h\times m \right)$, where $m$ and $h$ are much smaller than $n$, leading to a substantial reduction in the number of parameters. Besides, it offers a flexible trade-off in model capacity by adjusting $m$, which helps mitigate potential underfitting issues. 
Second, the reparameterization strategy improves the optimization process. Instead of only $\boldsymbol{x}_v$ participating in the optimization of $\boldsymbol{C}_{v:}$, it introduces a learnable transformation matrix $\boldsymbol{W}$ that adaptively estimates the node-dependent matrix $\boldsymbol{S}$, thereby elegantly solving the optimization challenge.
Finally, it establishes a connection between node-wise ensemble coefficients and view-wise ensemble coefficients. If only the node-agnostic matrix $\boldsymbol{T}$ is used, the node-wise ensemble coefficients will simply collapse into view-wise ensemble coefficients.

\subsection{Why do MGFNNs work?}
Here, we simply analyze the effectiveness of MGFNNs from an information-theoretic perspective. The objective of node classification is to learn a function $f$ on the rooted graph $\mathcal{G}^{[v]}$ with label $\boldsymbol{y}_v$ \cite{GNN_vs_GAMLP}. From the information-theoretic perspective, learning $f$ by minimizing cross-entropy loss is equivalent to maximizing the mutual information $I(\mathcal{G}^{[v]}; \boldsymbol{y}_i)$. If we treat $\mathcal{G}^{[v]}$ as a joint distribution of two random variables $\boldsymbol{X}^{[v]}$ and $\mathcal{E}^{[v]}$, which represent node features and edges in $\mathcal{G}^{[v]}$ respectively, we have: $I(\mathcal{G}^{[v]}; \boldsymbol{y}_v) = I(\boldsymbol{X}^{[v]}, \mathcal{E}^{[v]}; \boldsymbol{y}_v) =  I(\mathcal{E}^{[v]}; \boldsymbol{y}_v) + I(\boldsymbol{X}^{[v]}; \boldsymbol{y}_v \vert \mathcal{E}^{[v]})$. 
Here, $I(\mathcal{E}^{[v]}; \boldsymbol{y}_v)$ depends solely on the edges and labels, meaning that MLPs can only maximize $I(\boldsymbol{X}^{[v]}; \boldsymbol{y}_v \vert \mathcal{E}^{[v]})$. In the extreme case, $I(\boldsymbol{X}^{[v]}; \boldsymbol{y}_v \vert \mathcal{E}^{[v]})$ may be zero if $\boldsymbol{y}^{[v]}$ is conditionally independent of $\boldsymbol{X}^{[v]}$ given $\mathcal{E}^{[v]}$. For example, when each node is labeled by its degree or whether it forms a triangle. Then MLPs and MGFNNs would fail to learn meaningful functions. However, such scenarios are rare and unlikely in the practical settings relevant to our work. In real-world node classification tasks, node features and structural roles are often highly correlated \cite{lerique2020joint,GLNN}, allowing MLPs to achieve reasonable performance even when based solely on node features. Hence, MGFNNs have the potential to achieve much better results.

\section{Experiments}
In this section, we conduct a series of experiments to answer the following research questions(RQ): \textbf{RQ1}: How do MGFNNs compare to MLPs and MGNNs? \textbf{RQ2}: How efficient are MGFNNs compared to MGNNs? \textbf{RQ3}: How to visually explain the learned node-wise ensemble coefficients? \textbf{RQ4}: How does MGFNN+ compare to view-wise ensemble distillation methods? \textbf{RQ5}: How do different hyperparameters affect MGFNNs?

\subsection{Experimental Setup}
\subsubsection{Datasets}
We conduct experiments on 4 small datasets ACM, IMDB, IMDB5K, DBLP \cite{HAN,DMG}, and 2 large datasets ArXiv, MAG \cite{OGB}. Table \ref{Tab: Dataset Statistics} summarizes the dataset statistics.

\begin{table}[h]
\begin{center}
\caption{Dataset Statistics.}\label{Tab: Dataset Statistics}
\vspace{-0.5cm}
\begin{tabular}{l|c|c|c|c|c|c} 
\toprule
Dataset               & Nodes       & Views    & Edges & Feats & Train/Val/Test & Classes          \\ 
\midrule
\multirow{2}{*}{ACM}    & \multirow{2}{*}{3,025} & Paper-Subject-Paper & 2,210,761 & \multirow{2}{*}{1,870} & \multirow{2}{*}{600/300/2125} & \multirow{2}{*}{3}  \\
                        &                        & Paper-Author-Paper & 29,281    &                        &                               & \\
\midrule
\multirow{2}{*}{IMDB}   & \multirow{2}{*}{3,550} & Movie-Director-Movie & 13,788 & \multirow{2}{*}{2,000} & \multirow{2}{*}{300/300/2950} & \multirow{2}{*}{3}  \\
                        &                        & Movie-Actor-Movie    & 66,428 &                        &                               & \\
\midrule
\multirow{2}{*}{IMDB5K} & \multirow{2}{*}{4,780} & Movie-Director-Movie & 21,018 & \multirow{2}{*}{1,232} & \multirow{2}{*}{300/300/2687} & \multirow{2}{*}{3}  \\
                        &                        & Movie-Actor-Movie    & 98,010 &                        &                               & \\
\midrule
\multirow{2}{*}{DBLP}   & \multirow{2}{*}{7,907} & Paper-Paper-Paper & 94,677 & \multirow{2}{*}{2,000} & \multirow{2}{*}{80/200/7627} & \multirow{2}{*}{4}  \\
                        &                        & Paper-Author-Paper    & 144,783 &                        &                               & \\
\midrule
\multirow{2}{*}{ArXiv}  & \multirow{2}{*}{81,634}& Paper-Paper & 1,019,624 & \multirow{2}{*}{128} & \multirow{2}{*}{47084/18170/16380} & \multirow{2}{*}{40}  \\
                        &                        & Paper-Author-Paper & 1,985,544    &                        &                               & \\
\midrule
\multirow{2}{*}{MAG}  & \multirow{2}{*}{216,863}& Paper-Paper & 3,812,069 & \multirow{2}{*}{128} & \multirow{2}{*}{181517/19998/15348} & \multirow{2}{*}{10}  \\
                        &                        & Paper-Author-Paper & 10,663,501    &                        &                               & \\
\bottomrule
\end{tabular}
\vspace{-1cm}
\end{center}
\end{table}

\subsubsection{Teacher Models}
Similar to SAGE \cite{SAGE} used in GLNN \cite{GLNN}, we use RSAGE as the teacher. We also show the impact of other teacher models including RGCN \cite{RGCN}, RGAT \cite{GAT}, and HAN \cite{HAN} in Section \ref{Sec: Teacher Architecture}.


\subsubsection{Implementation Details}
All experiments are conducted on a 32GB NVIDIA Tesla V100 GPU. The models are trained with a learning rate of 0.01 and a weight decay selected from the set $\{1e-3, 5e-3, 5e-4, 0 \}$. For all datasets, the number of layers is set to 2, the hidden dimension for MGNNs is 128, and the hidden dimension for MLPs and MGFNNs is also 128 (1024 for ArXiv and MAG). The trade-off parameter $\lambda$ is set to 0, as non-zero values did not lead to significant improvements \cite{GLNN}. The low-rank parameter $m$ is searched from $\{ 1, 2, 3 \}$, and the regularization weight $\gamma$ is searched from $\{0.1, 0.01, 0.001\}$. 
Results are reported as the mean and standard deviation over five runs with different random seeds. Model performance is evaluated based on accuracy, and the model with the highest validation accuracy is selected for testing. Our implementation is available at \url{https://github.com/Cloudy1225/MGFNN}.

\subsubsection{Transductive vs. Inductive} \label{Sec: Transductive vs. Inductive}
To fully evaluate our model, we conduct node classification in two settings: transductive (\emph{tran}) and inductive (\emph{ind}). For \emph{tran}, we train models on $\mathcal{G}$, $\boldsymbol{X}^L$, and $\boldsymbol{Y}^L$, while evaluate them on $\boldsymbol{X}^U$ and $\boldsymbol{Y}^U$. During distillation, we generate soft labels for every node in the graph (i.e., $\boldsymbol{z}_v$ for $v \in \mathcal{V}$). For \emph{ind}, we follow GLNN \cite{GLNN} to randomly select out 20\% test data for inductive evaluation. Specifically, we separate the unlabeled nodes $\mathcal{V}^U$ into two disjoint subsets: observed $\mathcal{V}^U_{obs}$ and inductive $\mathcal{V}^U_{ind}$, leading to three separate graphs $\mathcal{G} = \mathcal{G}^L \sqcup \mathcal{G}_{obs}^U \sqcup \mathcal{G}_{ind}^U$ with no shared nodes. During training, the edges between $\mathcal{G}^L \sqcup \mathcal{G}_{obs}^U$ and $\mathcal{G}_{ind}^U$ are removed but are used during inference. Node features and labels are partitioned into three disjoint sets: $\boldsymbol{X} = \boldsymbol{X}^L \sqcup \boldsymbol{X}_{obs}^U \sqcup \boldsymbol{X}_{ind}^U$ and $\boldsymbol{Y} = \boldsymbol{Y}^L \sqcup \boldsymbol{Y}_{obs}^U \sqcup \boldsymbol{Y}_{ind}^U$. During distillation, soft labels are generated for nodes in the labeled and observed subsets i.e., $\boldsymbol{z}_v$ for $v \in \mathcal{V}^L \sqcup \mathcal{V}^U_{obs}$.

\subsection{Accuracy (RQ1)}\label{Sec: Accuracy}
We first compare MGFNNs with MLPs and MGNNs under the standard transductive setting. As shown in Table \ref{Tab: Transductive Setting}, both MGFNN and MGFNN+ show significant improvements over vanilla MLPs. Compared to MGNNs, MGFNN exhibits a slight performance degradation on 4/6 datasets, while MGFNN+ achieves the best performance on 5/6 datasets. On average, MGFNN+ improves performance by 1.64\% over MGFNN across different datasets, highlighting the effectiveness of our proposed multi-view ensemble distillation strategy.

\begin{table}[!ht]
    \begin{center}
    \setlength{\tabcolsep}{4pt}
    {\caption{Classificatiom accuracy under the transductive setting. $\Delta_1$, $\Delta_2$, $\Delta_3$ represents the difference between the MGFNN+ and MLP, RSAGE, MGFNN, respectively.}\label{Tab: Transductive Setting}}
    \vspace{-0.4cm}
    \begin{tabular}{l|ccc|cccc}
    \toprule
    Dataset & MLP & RSAGE & MGFNN & MGFNN+ & $\Delta_1$ & $\Delta_2$ & $\Delta_3$ \\ \midrule
    ACM & 67.77±1.40 & 87.92±0.28 & 87.48±1.85 & \textbf{89.10±0.50} & 21.33 & 1.18 & 1.62 \\
    IMDB & 57.76±1.88 & 63.86±0.22 & 64.56±0.35 & \textbf{65.95±0.55} & 8.19 & 2.09 & 1.39 \\
    IMDB5K & 49.86±0.59 & 58.21±0.70 & 59.23±0.58 & \textbf{60.04±0.28} & 10.18 & 1.83 & 0.81 \\
    DBLP & 57.10±0.29 & 72.35±1.29 & 71.79±1.50 & \textbf{74.18±2.54} & 17.08 & 1.83 & 2.39 \\
    ArXiv & 64.28±0.16 & 77.69±0.13 & 76.42±0.36 & \textbf{78.25±0.23} & 13.97 & 0.56 & 1.83 \\
    MAG & 52.13±0.85 & \textbf{62.74±0.45} & 58.50±0.87 & 60.32±0.64 & 8.19 & -2.42 & 1.82 \\ 
    \bottomrule
    \end{tabular}
    \vspace{-0.4cm}
    \end{center}
\end{table}

To fully evaluate the performance of MGFNNs, we further conduct experiments in a realistic production (\textit{prod}) setting that includes both inductive (\textit{ind}) and transductive (\textit{tran}) predictions, as detailed in Section \ref{Sec: Transductive vs. Inductive}. We report \textit{tran}, \textit{ind} results, and interpolated \textit{prod} results in Table \ref{Tab: Production Setting}. The \textit{prod} results provide a clearer understanding of the model's generalization and its accuracy in production environments.
In Table \ref{Tab: Production Setting}, we observe that MGFNN and MGFNN+ can still outperform MLPs by large margins. On 5/6 datasets, the MGFNN+ \textit{prod} results are competitive with those of MGNNs, suggesting that MGFNN+ can be deployed as a much faster model with no or only slight performance loss. 

However, on the ArXiv and MAG datasets, the MGFNN+ performance is lower than that of MGNNs. We hypothesize that this is due to these datasets having particularly challenging data splits, which lead to a distribution shift between test and training nodes. This shift makes it difficult for MGFNNs to capture the patterns without leveraging neighbor information, as MGNNs do. Nonetheless, it is important to note that MGFNNs consistently outperform vanilla MLPs.

\begin{table}[!ht]
    \begin{center}
    \setlength{\tabcolsep}{2.5pt}
    {\caption{Classificatiom accuracy under the production setting with both inductive and transductive predictions. \textit{ind} results on $\mathcal{V}_{ind}^U$, \textit{tran} results on $\mathcal{V}_{obs}^U$, and the interpolated \textit{prod} results are reported ($prod=0.2*ind+0.8*tran$).}\label{Tab: Production Setting}}
    \vspace{-0.4cm}
    \begin{tabular}{lc|ccc|cccc}
    \toprule
    Dataset & Eval & MLP & RSAGE & MGFNN & MGFNN+ & $\Delta_1$ & $\Delta_2$ & $\Delta_3$ \\ \midrule
    \multirow{3}{*}{ACM} & \textit{prod} & 67.77±1.40 & 87.19±1.35 & 85.82±1.34 & \textbf{87.59±1.64} & 19.82 & 0.40 & 1.77 \\
    ~ & \textit{ind} & 67.86±3.38 & 86.68±1.63 & 79.86±1.34 & 81.36±1.74 & 13.50 & -5.32 & 1.50 \\
    ~ & \textit{tran} & 67.75±1.06 & 87.32±1.40 & 87.31±1.37 & 89.14±1.69 & 21.39 & 1.82 & 1.83 \\ \midrule
    \multirow{3}{*}{IMDB} & \textit{prod} & 57.76±1.88 & 63.67±0.29 & 63.07±0.51 & \textbf{64.60±0.91} & 6.84 & 0.93 & 1.53 \\
    ~ & \textit{ind} & 57.12±2.50 & 63.80±1.46 & 59.08±1.95 & 59.93±1.66 & 2.81 & -3.87 & 0.85 \\
    ~ & \textit{tran} & 57.92±1.78 & 63.64±0.34 & 64.07±0.49 & 65.76±0.82 & 7.84 & 2.12 & 1.69 \\ \midrule
    \multirow{3}{*}{IMDB5K} & \textit{prod} & 49.86±0.59 & 56.78±1.23 & 55.56±1.56 & \textbf{57.50±0.87} & 7.64 & 0.72 & 1.94 \\
    ~ & \textit{ind} & 48.90±2.08 & 56.46±2.43 & 51.10±3.64 & 51.81±1.88 & 2.91 & -4.65 & 0.71 \\
    ~ & \textit{tran} & 50.10±0.97 & 56.86±1.16 & 56.68±1.50 & 58.92±0.72 & 8.82 & 2.06 & 2.24 \\ \midrule
    \multirow{3}{*}{DBLP} & \textit{prod} & 57.10±0.29 & 72.08±0.67 & 70.69±0.76 & \textbf{73.01±0.48} & 15.91 & 0.93 & 2.32 \\
    ~ & \textit{ind} & 57.05±0.89 & 72.42±1.51 & 67.17±1.09 & 67.67±0.83 & 10.62 & -4.75 & 0.50 \\
    ~ & \textit{tran} & 57.11±0.41 & 72.00±0.52 & 71.57±0.72 & 74.35±0.42 & 17.24 & 2.35 & 2.78 \\ \midrule
    \multirow{3}{*}{ArXiv} & \textit{prod} & 64.28±0.16 & \textbf{77.69±0.13} & 72.44±0.21 & 75.58±2.33 & 11.30 & -2.11 & 3.14 \\
    ~ & \textit{ind} & 64.00±0.75 & 77.99±0.66 & 66.14±1.11 & 66.75±0.59 & 2.75 & -11.24 & 0.61 \\
    ~ & \textit{tran} & 64.34±0.35 & 77.62±0.24 & 74.02±0.12 & 77.79±2.91 & 13.45 & 0.17 & 3.77 \\ \midrule
    \multirow{3}{*}{MAG} & \textit{prod} & 52.13±0.85 & \textbf{62.65±0.57} & 56.65±1.24 & 58.59±1.87 & 6.46 & -4.06 & 1.94 \\
    ~ & \textit{ind} & 51.60±0.76 & 62.93±0.92 & 54.88±0.57 & 55.79±1.19 & 4.19 & -7.14 & 0.91 \\
    ~ & \textit{tran} & 52.26±0.91 & 62.58±0.65 & 57.10±1.51 & 59.29±2.05 & 7.03 & -3.29 & 2.19 \\
    \bottomrule
    \end{tabular}
    \end{center}
\end{table}

\subsection{Efficiency (RQ2)}\label{Sec: Efficiency}
Inference efficiency and accuracy are two key metrics for evaluating machine learning systems. With the increasing demand for graph learning applications in industry, there is a growing need for models that can perform inference with low latency. Here, we compare the inference times of RSAGE, RSAGE with neighbor sampling (NS), and MGFNNs on 10 randomly selected nodes. NS-10 indicates that each node receives messages from 10 sampled neighbors per edge type during inference. As shown in Table \ref{Tab: Inference Time}, our MGFNNs significantly outperform the baseline methods, achieving speedups ranging from 35.40× to 89.14× over the teacher RSAGE. This improvement is attributed to the fact that MGFNNs, which are essentially well-trained MLPs, eliminate the extensive multiplication-and-accumulation operations over the features of numerous neighbors in MGNNs. These results highlight the superior inference efficiency of our MGFNNs, underscoring their suitability for latency-sensitive deployments.

\begin{table}[!ht]
\begin{center}
\setlength{\tabcolsep}{8pt}
{\caption{Inductive inference time (in ms) on 10 randomly chosen nodes. NS-10 means inference neighbor sampling with fan-out 10 for each edge type.}\label{Tab: Inference Time}}
\vspace{-0.4cm}
\begin{tabular}{lcccccc} 
\toprule
Method & ACM & IMDB & IMDB5K & DBLP & ArXiv & MAG \\ 
\midrule
RSAGE & 9.125 & 6.463 & 6.568 & 6.443 & 12.329 & 15.510 \\
\midrule
\multirow{2}{*}{NS-10}  & 6.517 & 6.406 & 6.351 & 6.307 & 6.358 & 6.415 \\
                        & 1.40× & 1.01× & 1.03× & 1.02× & 1.94× & 2.42× \\
\midrule
\multirow{2}{*}{MGFNNs} & 0.182 & 0.181 & 0.179 & 0.182 & 0.176 & 0.174  \\
                        & \textbf{50.14×} & \textbf{35.71×} & \textbf{36.69×} & \textbf{35.40×} & \textbf{70.05×} & \textbf{89.14×} \\
\bottomrule
\end{tabular}
\vspace{-0.4cm}
\end{center}
\end{table}

\subsection{Interpretability (RQ3)}\label{Sec: Interpretability}
\begin{figure}[h]
    \centering
    \vspace{-0.5cm}
    \subfigure[ACM]{\includegraphics[width=0.325\linewidth]{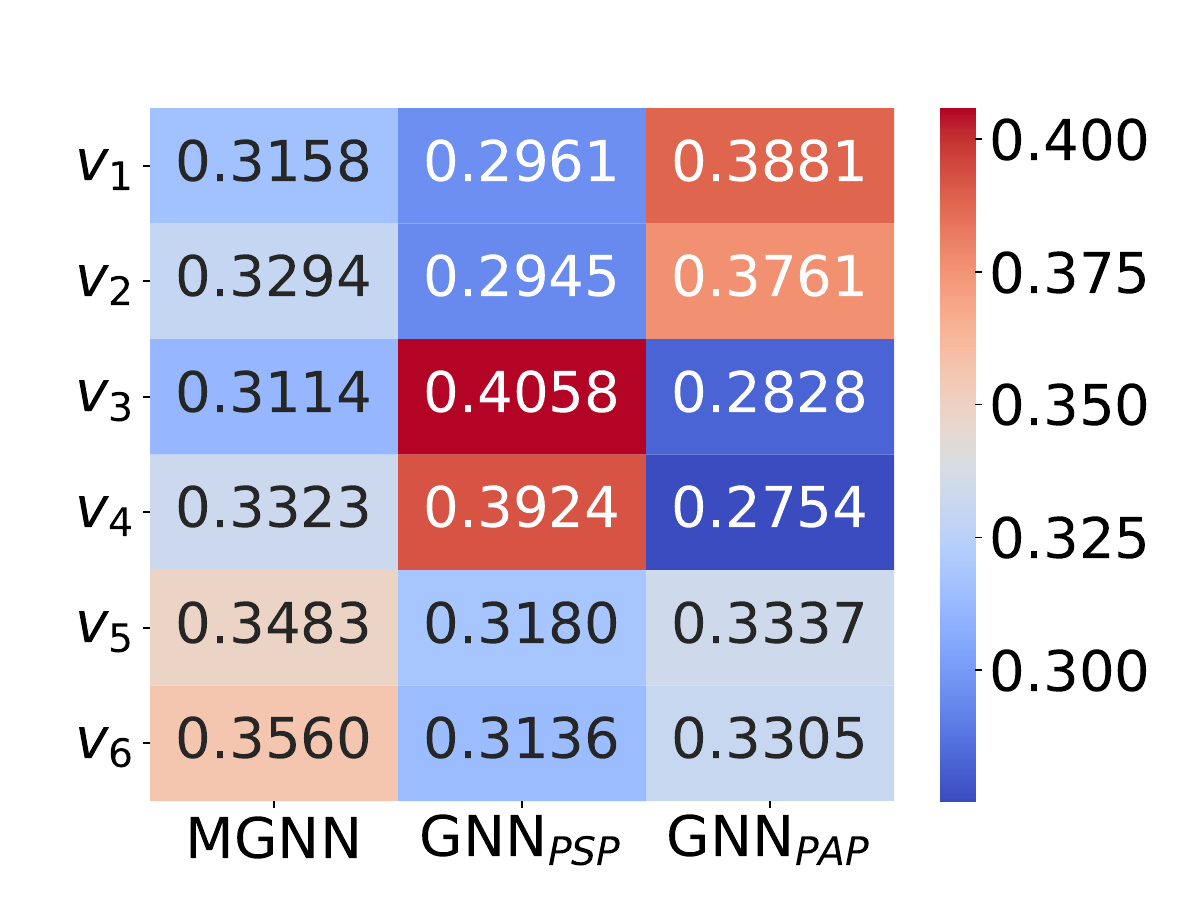}}  
    \subfigure[IMDB]{\includegraphics[width=0.325\linewidth]{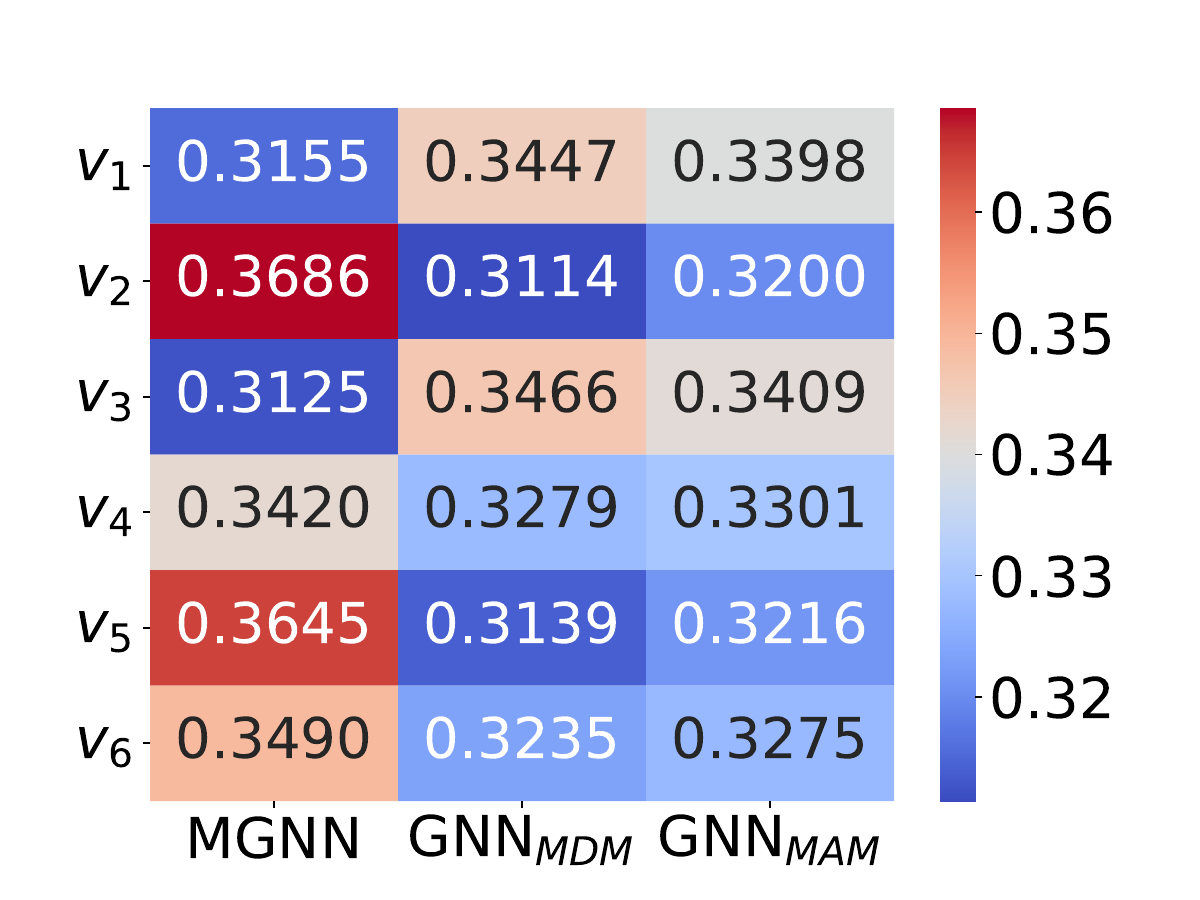}}
    \subfigure[MAG]{\includegraphics[width=0.325\linewidth]{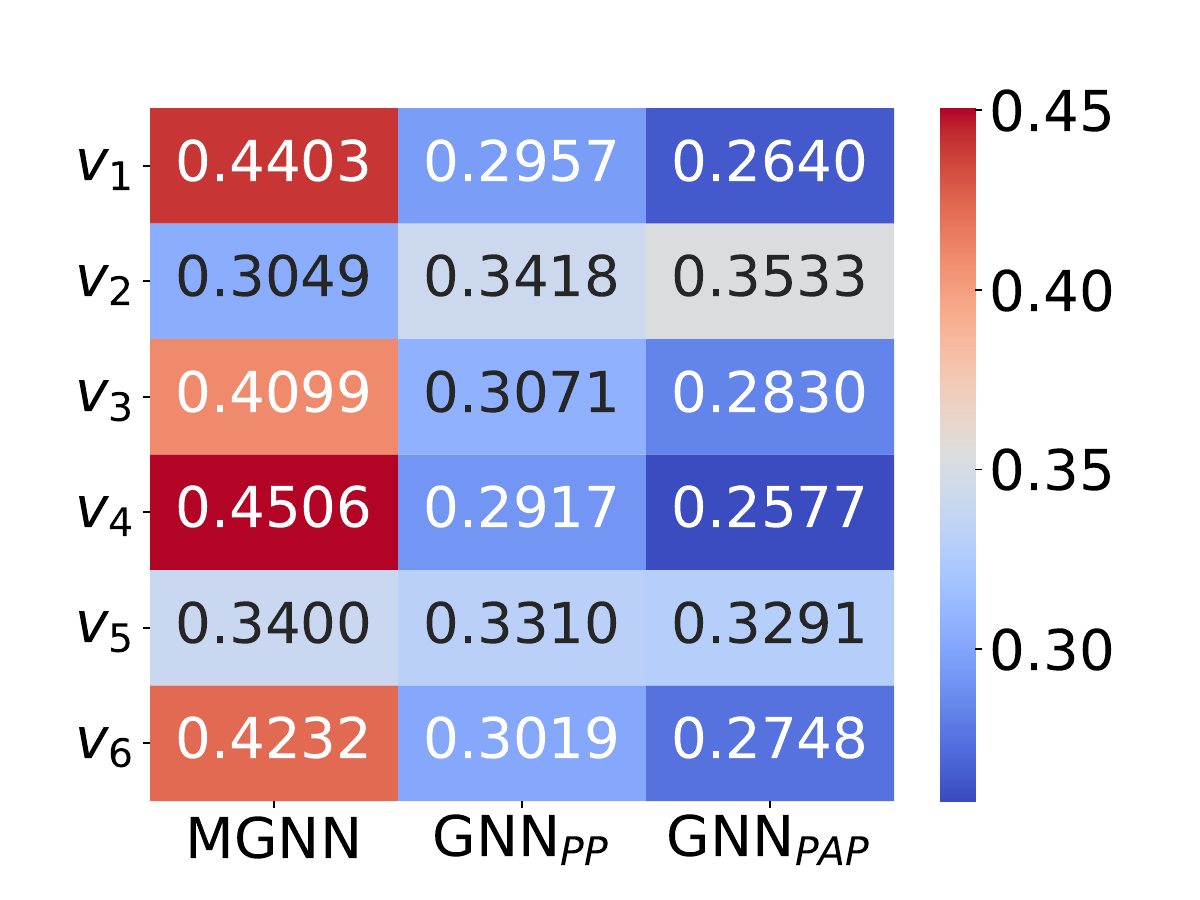}}
    \vspace{-0.2cm}
    \caption{Visualization of learned node-wise ensemble coefficients for 6 randomly selected nodes on ACM, IMDB, and MAG.}
    \vspace{-0.5cm}
    \label{Fig: Coef HeatMap}
\end{figure}

To gain visual insights into MGFNN+, we present heatmaps that depict the learned node-wise ensemble coefficients for six randomly selected nodes across the ACM, IMDB, and MAG datasets. As shown in Figure \ref{Fig: Coef HeatMap}, MGFNN+ assigns personalized ensemble coefficients to different nodes. For instance, on the MAG dataset, where MGNN, GNN$_{PSP}$, and GNN$_{PAP}$ encode the Whole, PSP, and PAP views, respectively, we observe the following patterns: 1) No Clear Dominance: No single view is consistently the most important across all nodes. The coefficients for each view are interspersed, suggesting that all views contribute significantly to the ensemble for different nodes. 2) Variation in Importance: The coefficients vary across nodes, indicating that the importance of each view differs for different nodes. For example, for node $v_1$, the PAP view has the highest coefficient (0.3881), while for node $v_3$, the PSP view holds the highest (0.4058).

\subsection{Ablation Studies (RQ4)}
To show the superiority of our node-wise multi-view ensemble distillation, we substitute node-wise ensemble coefficients in Eq.~\eqref{Eq: Node-wise MVED} with view-wise ensemble coefficients in Eq.~\eqref{Eq: View-wise MVED}. These coefficients can be uniformly set to $1/(r+1)$ (MEAN), defined as $(r+1)$-dimensional learnable parameters (PARA), or computed adaptively using attention (ATTN) \cite{MTAAM} or gradient (GRAD) \cite{AEKD}.
As reported in Table \ref{Tab: Ablation Study}, these view-wise ensemble approaches consistently outperform MGFNN in most cases, indicating that view-specific GNNs provide additional semantic knowledge beneficial for classification. However, they all yield inferior classification performance compared to MGFNN+, emphasizing the superior performance of our node-wise ensemble strategy.

\begin{table}[h]
\begin{center}
\setlength{\tabcolsep}{4.5pt}
{\caption{Ablation study on node/view-wise ensemble distillation.}\label{Tab: Ablation Study}}
\vspace{-0.5cm}
\begin{tabular}{lcccccc} 
\toprule
Method & ACM & IMDB & IMDB5K & DBLP & ArXiv & MAG \\ 
\midrule
MGFNN & 87.5±1.9 & 64.6±0.3 & 59.2±0.6 & 71.8±1.5 & 76.4±0.4 & 58.5±0.9 \\
MEAN & 88.3±0.9 & 64.5±0.5 & 59.2±0.7 & 71.8±1.5 & 75.3±0.2 & 58.7±0.8 \\
PARA & 88.3±1.1 & 64.6±0.1 & 59.3±0.4 & 71.8±1.7 & 77.2±0.2 & 59.9±0.7 \\
ATTN & 88.2±1.0 & 64.6±0.3 & 59.3±0.4 & 71.7±1.6 & 76.6±0.1 & 59.9±0.3 \\
GRAD & 87.8±1.1 & 64.3±1.0 & 59.1±0.7 & 72.0±2.0 & 75.4±0.4 & 59.4±0.7 \\
MGFNN+ & \textbf{89.1±0.5} & \textbf{66.0±0.6} & \textbf{60.0±0.3} & \textbf{74.2±2.5} & \textbf{78.2±0.2} & \textbf{60.3±0.6} \\
\bottomrule
\end{tabular}
\vspace{-0.6cm}
\end{center}
\end{table}


\subsection{Hyperparameter Analysis (RQ5)}

\subsubsection{\textbf{Teacher Architecture}}\label{Sec: Teacher Architecture}
\begin{figure}[h]
    \centering
    \vspace{-0.4cm}
    \subfigure[ACM]{\includegraphics[width=0.325\linewidth]{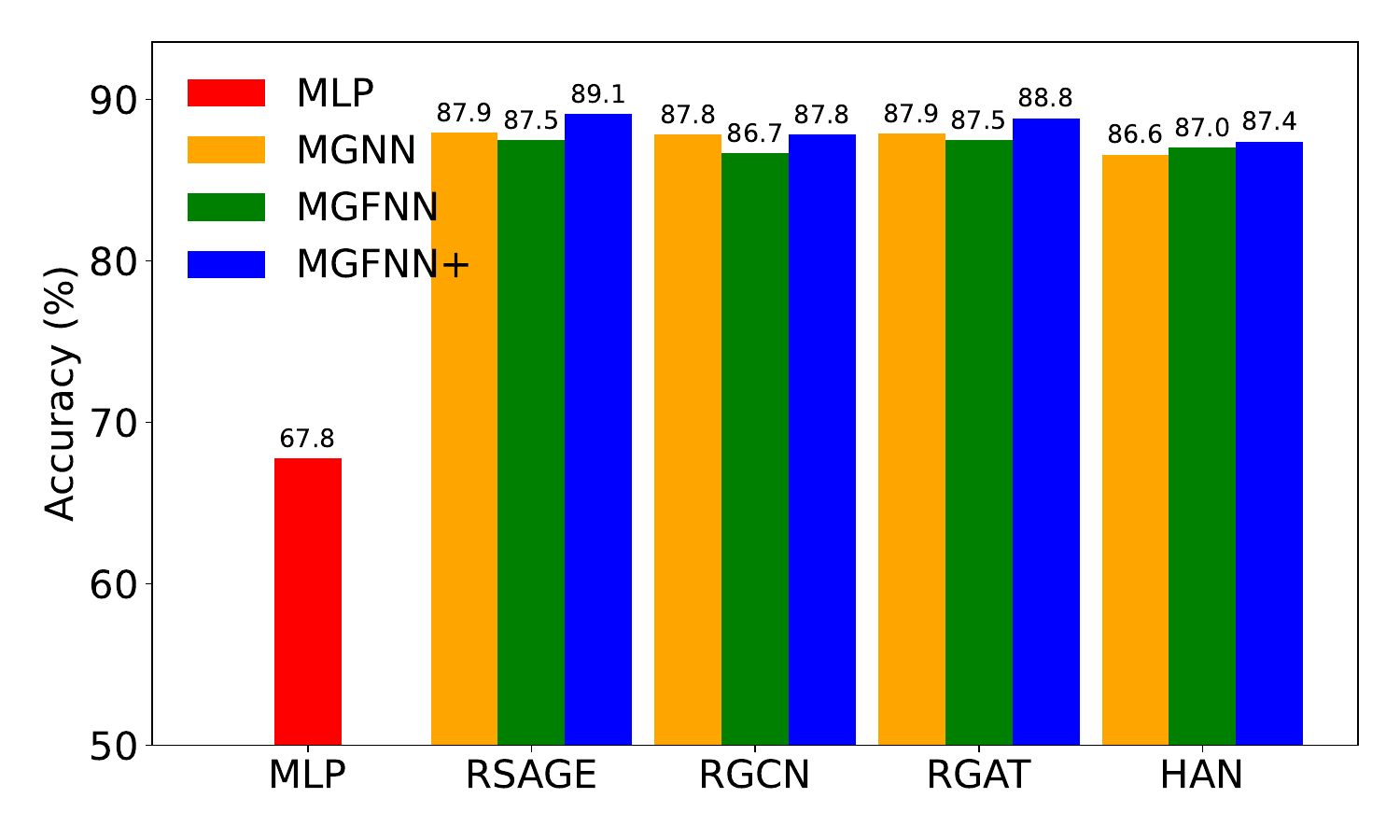}}  
    \subfigure[IMDB]{\includegraphics[width=0.325\linewidth]{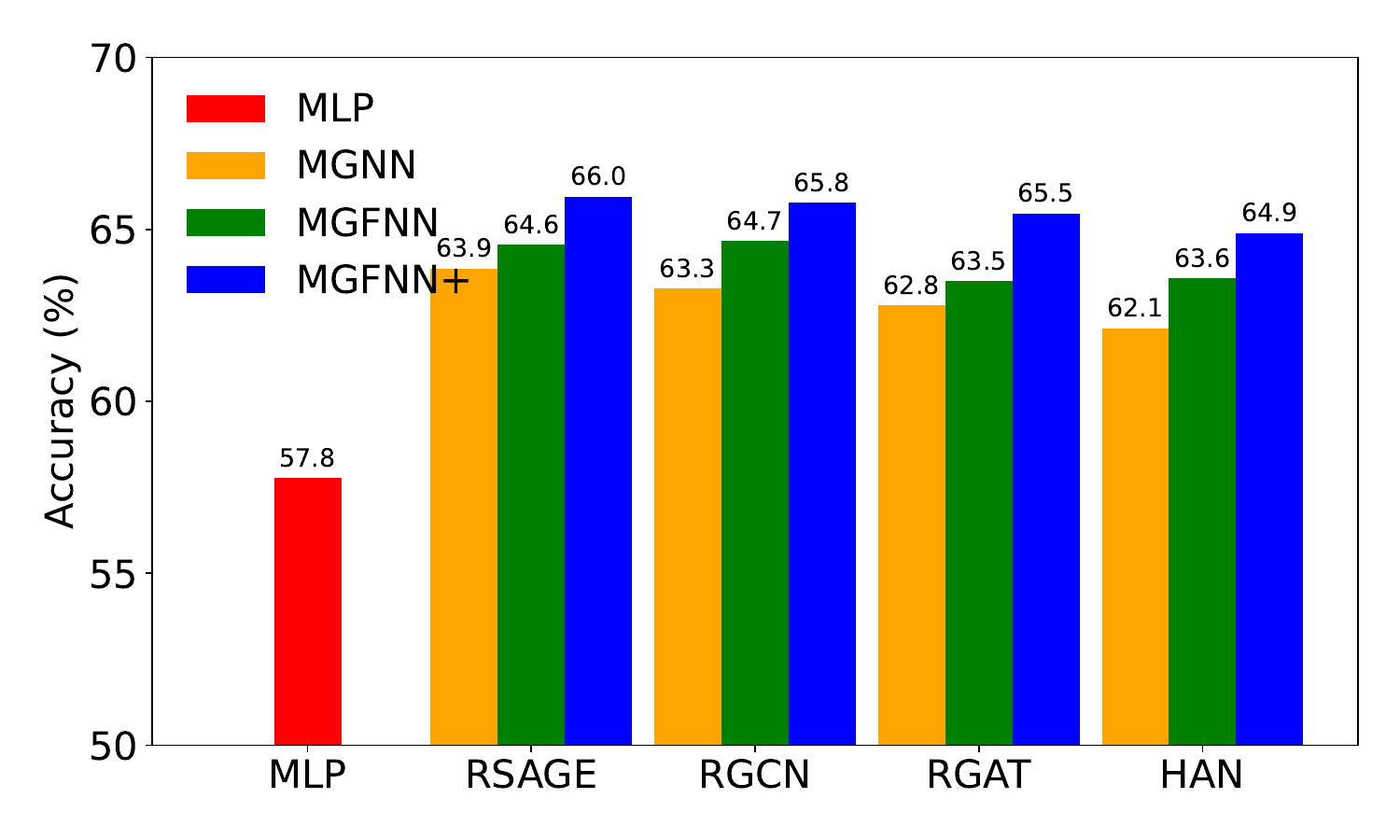}}
    \subfigure[ArXiv]{\includegraphics[width=0.325\linewidth]{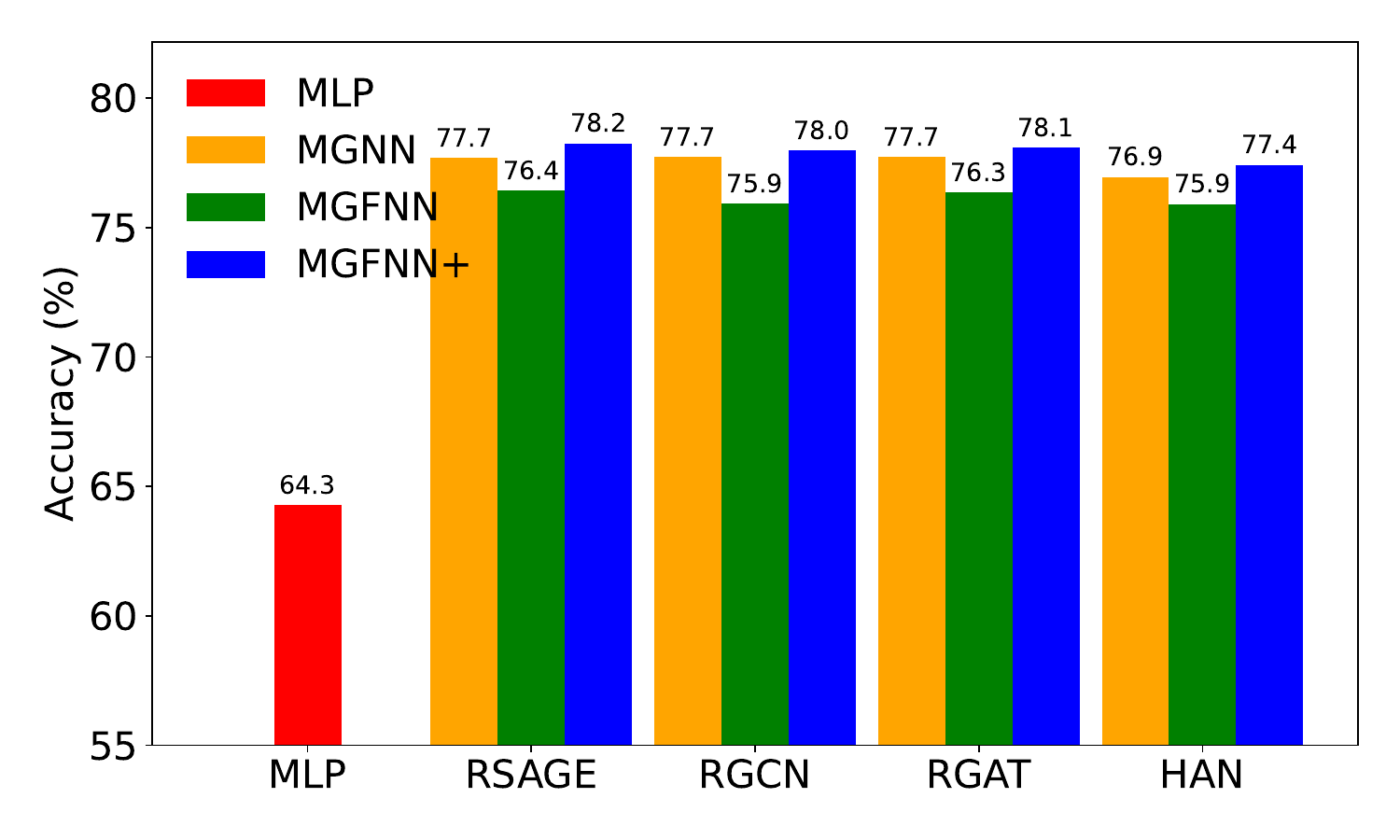}}
    \vspace{-0.2cm}
    \caption{Transductive Accuracy vs. Teacher MGNN Architectures. MGFNNs can learn from different MGNN teachers to improve over MLPs and achieve comparable results.}
    \label{Fig: Param Teacher Architecture}
    \vspace{-0.4cm}
\end{figure}

We investigate whether MGFNNs can perform well when trained with different MGNNs.
In Figure \ref{Fig: Param Teacher Architecture}, we present the transductive performance of MGFNNs when distilled from RSAGE, RGCN, RGAT, and HAN, across ACM, IMDB, and ArXiv datasets. We see that MGFNNs can effectively learn from different teachers and outperform vanilla MLPs. MGFNN achieves comparable performance to teachers, while MGFNN+ consistently surpasses them, underscoring the efficacy of our proposed model.

\subsubsection{\textbf{Inductive Split Rate}}\label{Sec: Inductive Split Rate}
\begin{figure}[h]
    \centering
    \vspace{-0.4cm}
    \subfigure[ACM]{\includegraphics[width=0.325\linewidth]{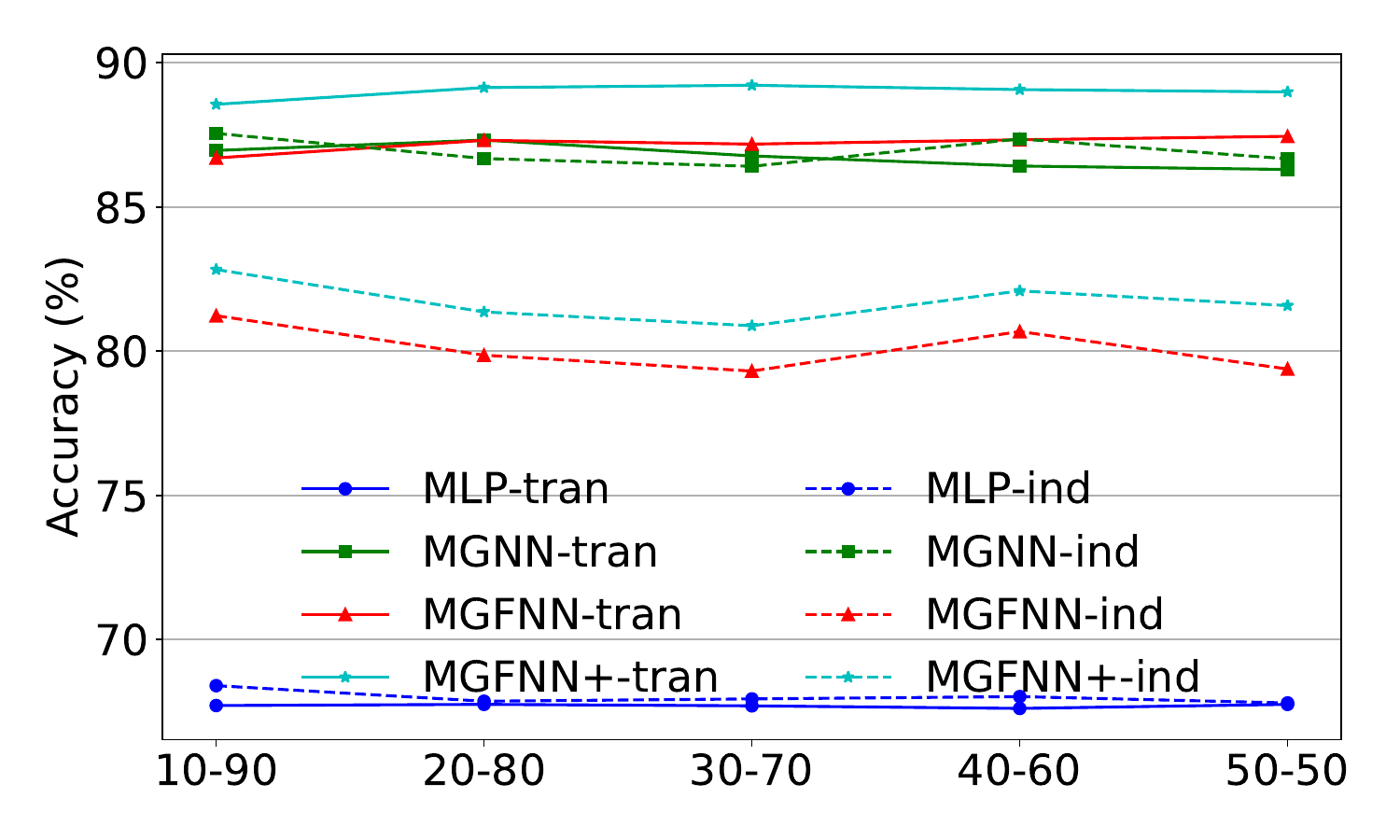}}  
    \subfigure[IMDB]{\includegraphics[width=0.325\linewidth]{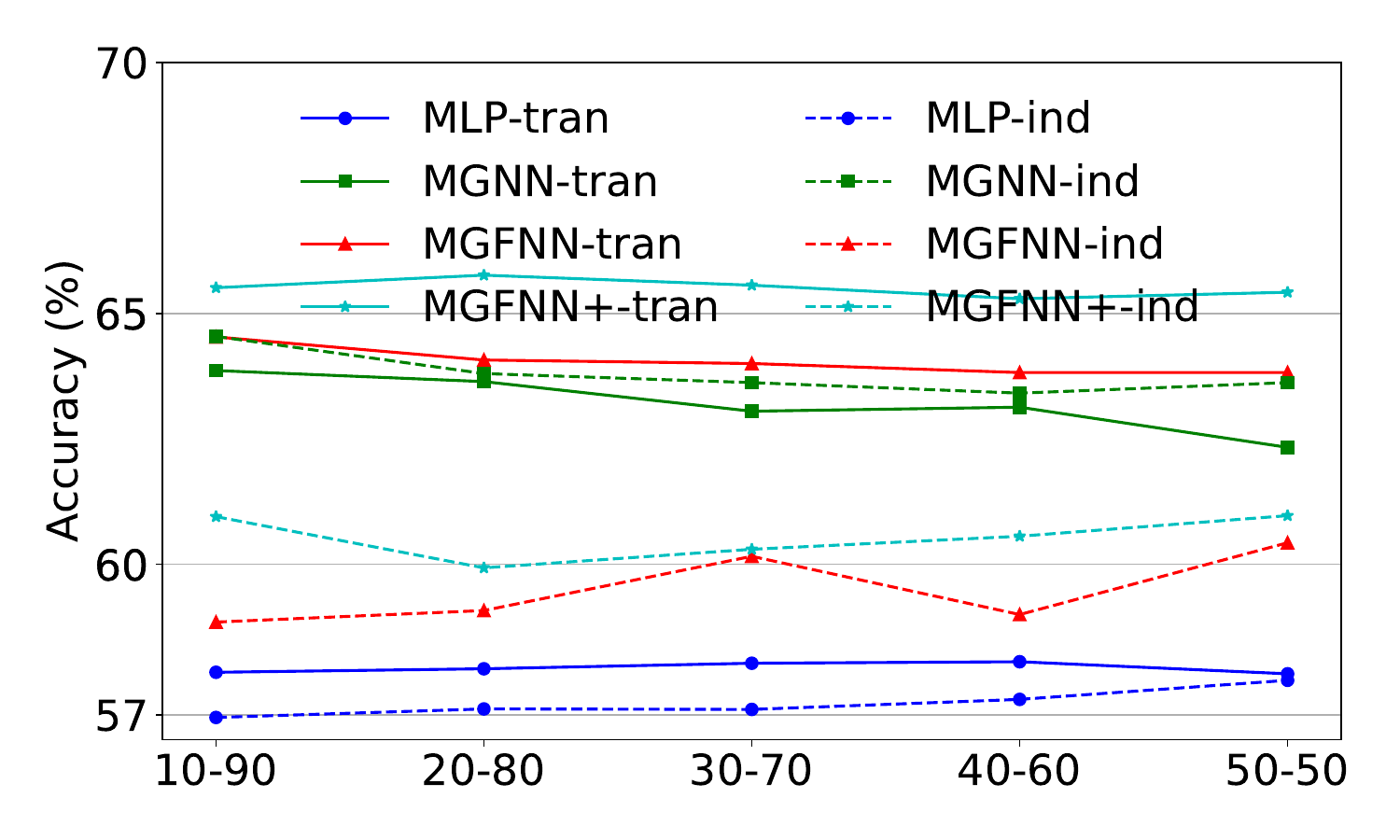}}
    \subfigure[ArXiv]{\includegraphics[width=0.325\linewidth]{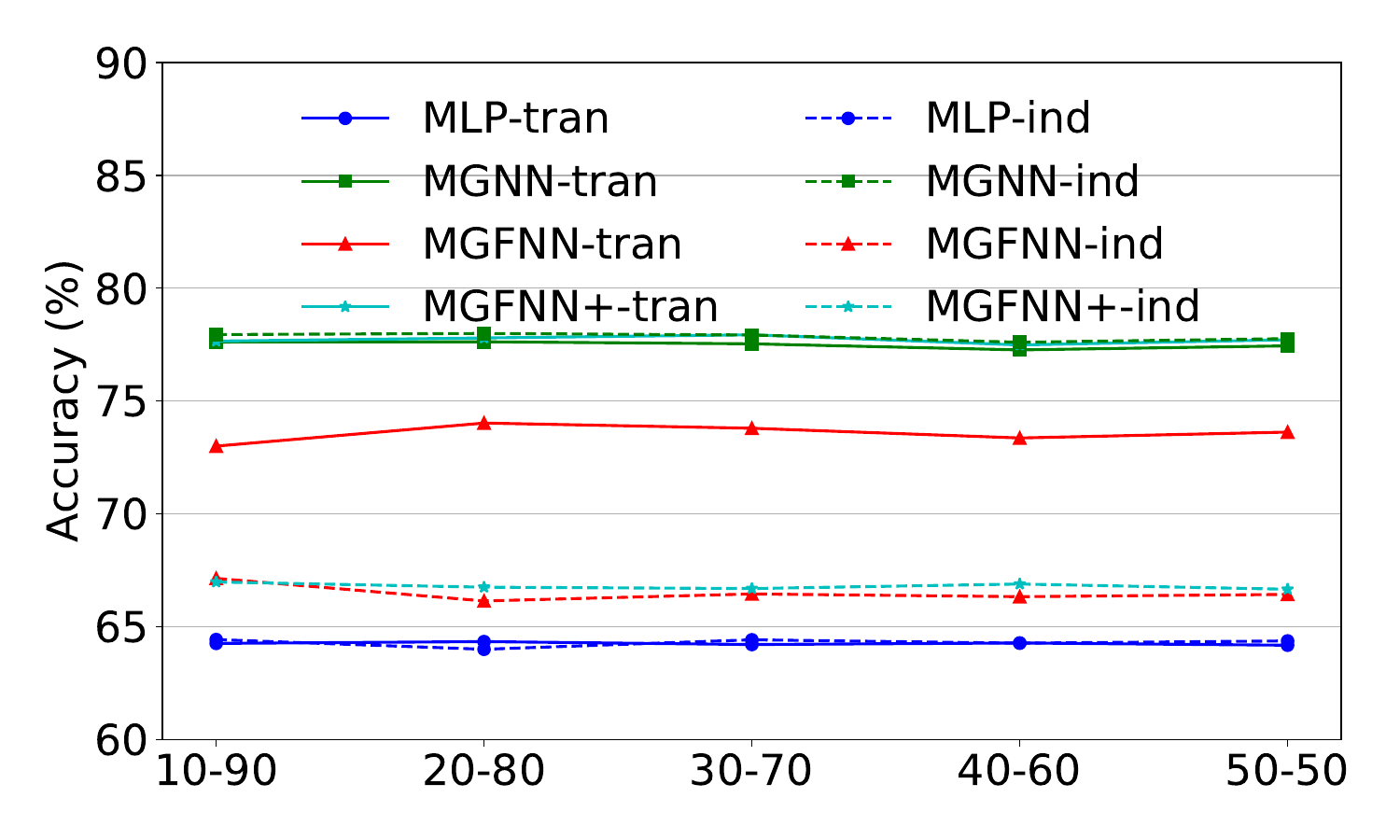}}
    \vspace{-0.2cm}
    \caption{Accuracy vs. Inductive:Transductive Ratio under the production setting.}
    \label{Fig: Param Split Rate}
    \vspace{-0.6cm}
\end{figure}

In Table \ref{Tab: Production Setting}, we employ a 20-80 split of the test data for inductive evaluation. 
Here, we conduct an ablation study on the inductive split rate under the production setting across ACM, IMDB, and ArXiv datasets. 
Figure \ref{Fig: Param Split Rate} shows that altering the inductive:transductive ratio in the production setting does not affect the accuracy much. 
We only consider rates up to 50-50 since having 50\% or more inductive nodes is exceedingly rare in practical scenarios. In cases where a substantial influx of new data occurs, practitioners can choose to retrain the model on the entire dataset before deployment.

\subsubsection{Other Hyperparameters}\label{Sec: Other Hyperparameters}
\begin{figure}[h]
    \centering
    \vspace{-0.4cm}
    \subfigure[Feature Noise Ratio]{\includegraphics[width=0.325\linewidth]{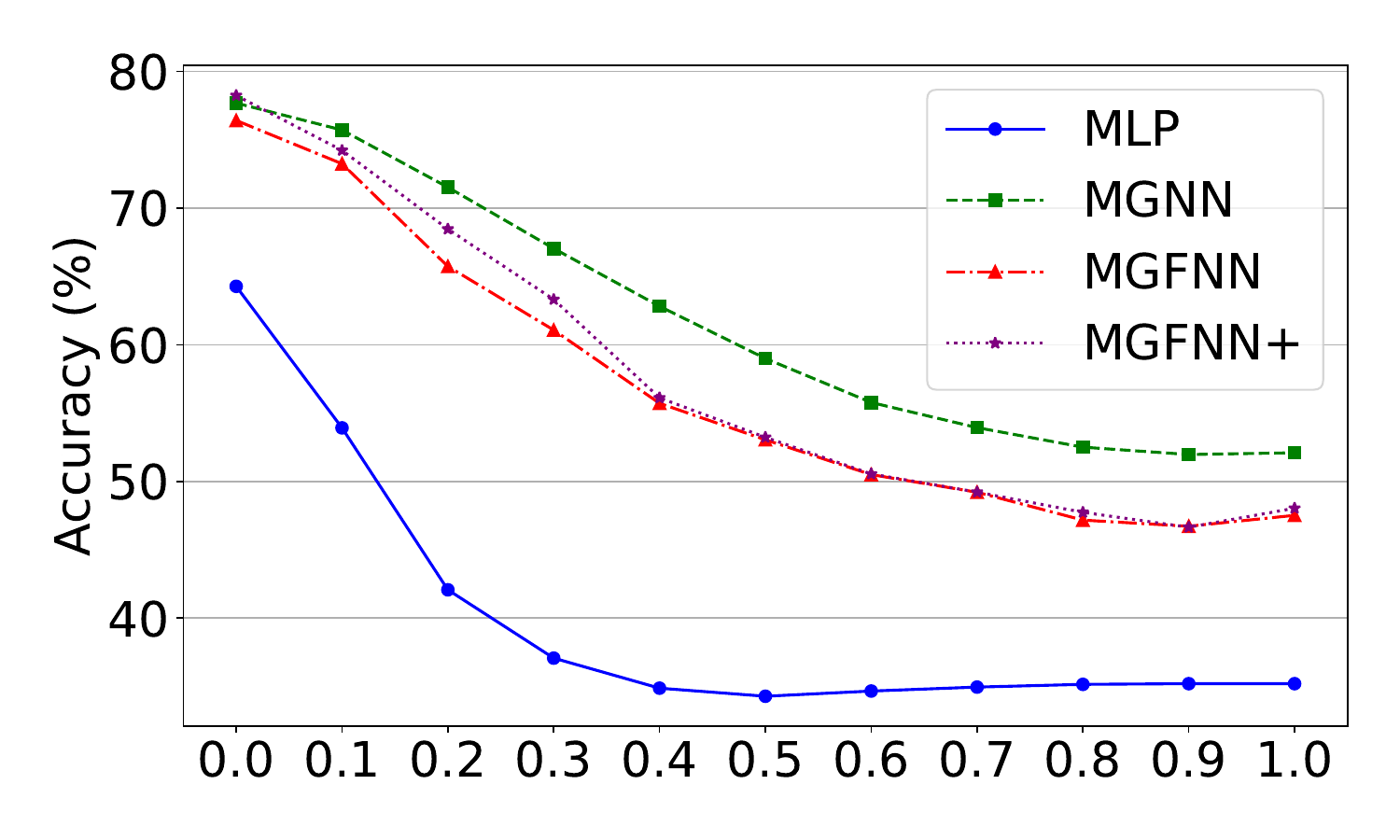}\label{Fig: Param Feature Noise}}
    \subfigure[Hidden Dimension]{\includegraphics[width=0.325\linewidth]{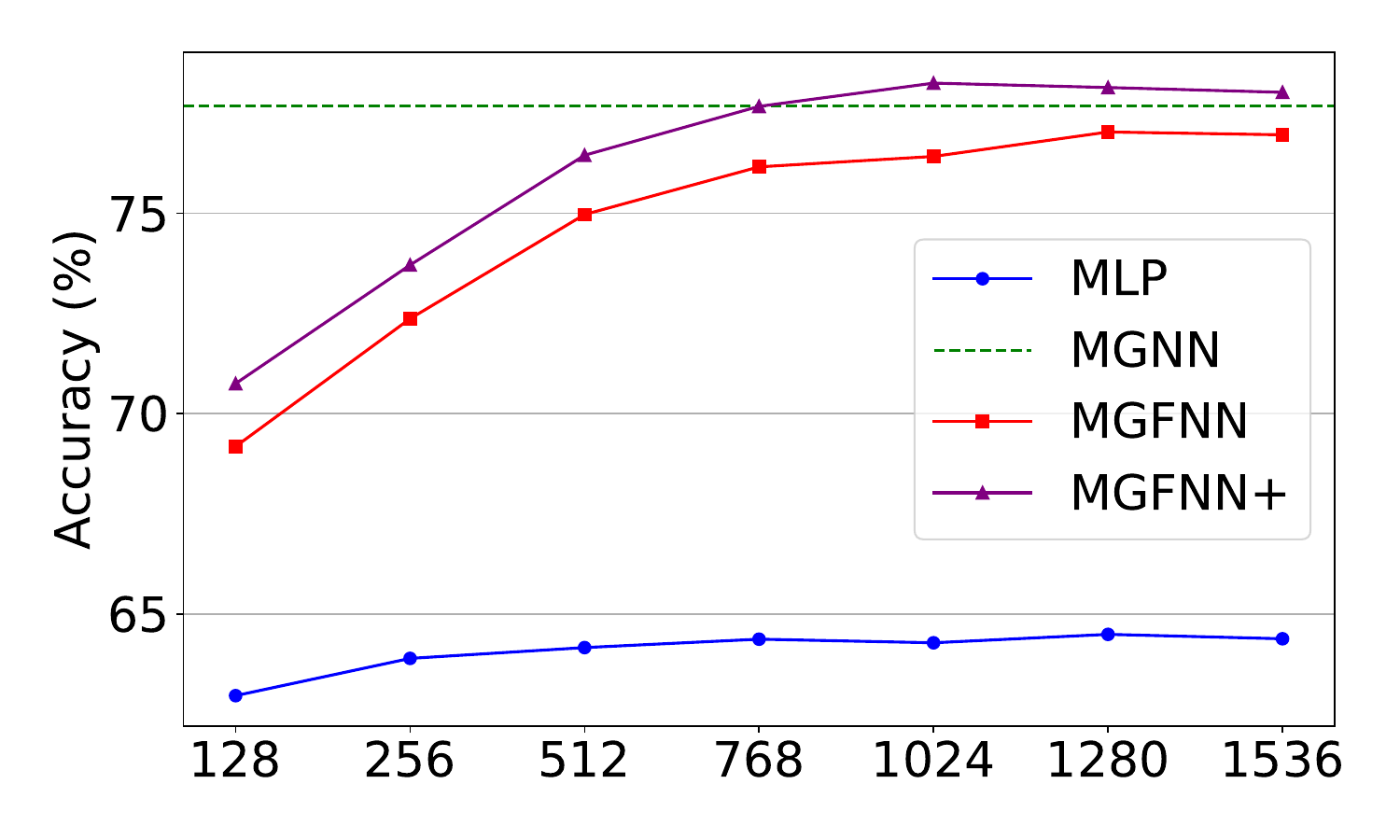}\label{Fig: Param Hidden Dimension}}
    \subfigure[Trade-off $\lambda$]{\includegraphics[width=0.325\linewidth]{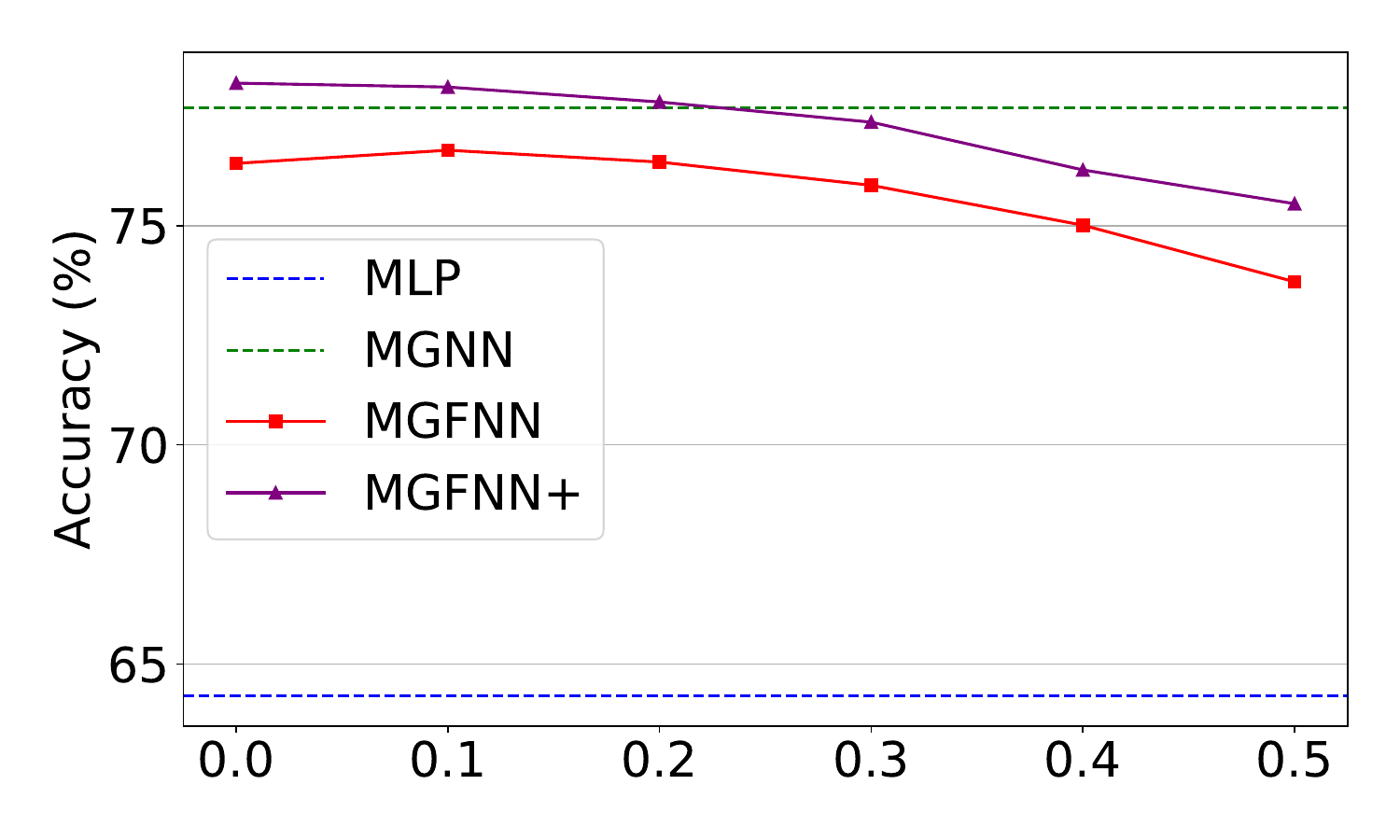}\label{Fig: Param Trade-off}}
    \vspace{-0.2cm}
    \caption{Accuracy vs. Feature Noise Ratio, Hidden Dimension, and $\lambda$ on ArXiv.}
    \label{Fig: Param on ArXiv}
    \vspace{-0.2cm}
\end{figure}

We further study the sensitivity of MGFNNs to noisy node features, the hidden dimension of MLPs, and the trade-off weight $\lambda$ on ArXiv.
Figure \ref{Fig: Param Feature Noise} shows that as the noise level increases, the accuracy of MLPs and MGFNNs declines faster than MGNNs, while the performance of MGFNNs and MGNNs remains comparable at lower noise levels. 
Figure \ref{Fig: Param Hidden Dimension} shows that as the hidden dimension increases, the performance of MGFNNs first increases and then stabilizes, with the performance gap between MGFNNs and MLPs widening. 
Figure \ref{Fig: Param Trade-off} shows that non-zero values of $\lambda$ are not very helpful, which is also observed in GLNN \cite{GLNN}.

\section{Conclusion}
In this paper, we propose MGFNN and MGFNN+ to combine both MGNNs' superior performance and MLPs' efficient inference. MGFNN directly trains student MLPs with node features as input and soft labels from teacher MGNNs as targets, and MGFNN+ further distills multiplex semantic knowledge into student MLPs through the multi-view ensemble distillation. We develop a low-rank approximation-based parameterization technique to learn node-wise coefficients, enabling adaptive knowledge ensemble for different nodes. Extensive experiments on six multiplex graph datasets show the accuracy, efficiency, and interpretability of MGFNNs, highlighting their potential for latency-sensitive applications.



\begin{credits}
\subsubsection{\ackname}
This work is partially supported by the National Natural Science Foundation of China (62306137, 62272469), and the Science and Technology Innovation Program of Hunan Province (2023RC1007).

\end{credits}
%
%
%
\bibliographystyle{splncs04}
\bibliography{main}
%




\end{document}